\documentclass{article}
\usepackage{setspace}
\usepackage{indentfirst}
\usepackage{amsfonts}
\usepackage{float}
\usepackage{subfigure}
\usepackage{graphicx}
\usepackage{bm}
\usepackage{array}
\usepackage[english]{babel}
\usepackage[letterpaper,top=2cm,bottom=2cm,left=2.5cm,right=2.5cm,marginparwidth=1.75cm]{geometry}
\usepackage{amsmath}
\usepackage{graphicx}
\usepackage{verbatim} 
\usepackage[colorlinks=true, allcolors=blue]{hyperref}
\usepackage{authblk}
\usepackage{comment}
\usepackage{multirow} 
\usepackage{multicol} 
\usepackage{colortbl}
\usepackage{arydshln}
\usepackage{booktabs}
\usepackage{makecell}

\usepackage{xcolor}

\title{Multi-spatial Multi-temporal Air Quality Forecasting with Integrated Monitoring and Reanalysis Data}

\author[a,b]{Yuxiao Hu}
\author[b,c]{Qian Li}
\author[d]{Xiaodan Shi}
\author[a]{Jinyue Yan}
\author[b]{Yuntian Chen\thanks{Corresponding author: ychen@eitech.edu.cn}}
\affil[a]{The Hong Kong Polytechnic University, Hong Kong, China}
\affil[b]{Ningbo Institute of Digital Twin, Eastern Institute of Technology, Ningbo, China;}
\affil[c]{Shanghai Jiao Tong University, Shanghai, China;}
\affil[d]{School of Business, Society and Technology, Mälardalens University, 72123 Västerås, Sweden}

\date{}
\begin{document}
\maketitle

\begin{abstract}
\begin{spacing}{1.2}

Accurate air quality forecasting is of paramount importance in the domains of public health, environmental monitoring and protection, and urban planning. However, existing methods often fail to effectively utilize information across different scales (varying spatial distances or temporal periods). Spatially, previous methods struggle to integrate information between individual monitoring stations and the overall city-scale, lacking flexibility in their interactions. Temporally, existing techniques often overlook or do not fully consider the periodic nature of variations in air quality, thus disregarding valuable insights across different time scales. To address these limitations, we present a novel \textbf{M}ulti-spatial \textbf{M}ulti-temporal air quality forecasting method based on \textbf{G}raph Convolutional Networks and \textbf{G}ated Recurrent Units (M2G2), bridging the gap in air quality forecasting across spatial and temporal scales. The proposed framework consists of two modules: Multi-scale Spatial GCN (MS-GCN) for spatial information fusion and Multi-scale Temporal GRU(MT-GRU) for temporal information integration. In the spatial dimension, the MS-GCN module employs a bidirectional learnable structure and a residual structure, enabling comprehensive information exchange between individual monitoring stations and the city-scale graph. Regarding the temporal dimension, the MT-GRU module adaptively combines information from different temporal scales through parallel hidden states. Leveraging meteorological indicators and four air quality indicators, we present comprehensive comparative analyses and ablation experiments, showcasing the higher accuracy of M2G2 in comparison to nine currently available advanced approaches across all aspects. The improvements of M2G2 over the second-best method on MAE and RMSE are as follows: PM2.5: (6.22\%, 6.63\%, 9.71\%) and (7.72\%, 6.67\%, 10.45\%), ${\rm PM}_{10}$: (5.78\%, 5.52\%, 8.26\%) and (6.43\%, 5.68\%, 7.73\%, ${\rm NO}_2$: (5.40\%, 9.73\%, 19.45\%) and (5.07\%, 7.76\%, 16.60\%), ${\rm O}_3$: (7.61\%, 7.17\%, 10.37\%) and (6.46\%, 6.86\%, 9.79\%). Furthermore, we demonstrate the effectiveness of each module of M2G2 by ablation study. Our proposed approach not only addresses the limitations of existing methods but also showcases its potential for advancing air quality forecasting using deep learning techniques.

\end{spacing}
\end{abstract}

\begin{spacing}{1.2}
\section{Introduction}

Air pollution poses a significant global public health risk, with air particles smaller than 2.5 micrometers in diameter, known as ${\rm PM}{2.5}$, capable of deeply penetrating the human lungs and bloodstream \cite{aerial-ganji2020predicting, street-level-feldman2023urban, survey-liu2022data}. This particulate matter is responsible for triggering a range of cardiovascular, respiratory, and other diseases. Accurately predicting ${\rm PM}{2.5}$ concentration and understanding its characteristics can have profound implications for various aspects of society. It can provide invaluable insights for public health officials, enabling them to develop effective strategies to improve air quality, such as implementing vehicle restrictions and regulating the siting of chemical plants. Additionally, accurate ${\rm PM}{2.5}$ prediction models hold the potential to guide individuals in making informed decisions regarding their daily activities, thereby safeguarding their well-being. Hence, the development of robust prediction models for ${\rm PM}{2.5}$ is an urgent and crucial task with far-reaching implications \cite{street-level-xu2022prediction, satellite-xiao2018ensemble, space–time-sun2021spatial}.

There are two key aspects that determine the accuracy of ${\rm PM}_{2.5}$ concentration prediction. Firstly, we need to consider the factors affecting ${\rm PM}_{2.5}$ concentrations as comprehensively as possible, such as wind speed, elevation, etc. which provide important prerequisites for the accuracy of predictions. Secondly, we need to adopt a reasonable information interaction and information fusion to combine the factors organically, since each influencing factor has a strong correlation. In particular, both information in the temporal and spatial dimensions needs to be considered.

Since the significant social and environmental value of ${\rm PM}_{2.5}$ concentrations prediction task, many methods have been proposed to solve this problem, which can be divided into knowledge-based and data-driven methods. Knowledge-based methods always rely on a large amount of prior knowledge to support the final decisions. Such as \cite{mathur08,chuang11} study the properties of transformation and diffusion of multiple pollutants, and provide air pollution prediction models through the prior knowledge of physical–chemical processes. Since ${\rm PM}_{2.5}$ concentration is influenced by a large number of factors and there are complex nonlinear relationships and stochasticity among the factors, knowledge-based methods often have poor flexibility because of heavy reliance on domain knowledge. Data-driven methods often use historical data to train the proposed methods, so that the models capture the potential connections in the data to predict future demand. 

Classical data-driven methods include statistical methods and machine learning algorithms. Statistical methods \cite{briggs07,wang09,rekhi20} always require a predetermined function between inputs and predicted value, which is not friendly to complex systems and may not be effective at capturing implicit relationships in long-term air quality prediction tasks. 
Some machine learning methods such as support vector regression (SVR) \cite{svr}, artificial neural networks (ANNs) \cite{ann} and random forest algorithm \cite{yu16} introduce nonlinear structures to improve the representation of complex systems. However, these methods ignore the correlation of data between different regions. 

Deep learning as a revolutionary data-driven model has been successfully applied in a wide range of fields, including computer vision, natural language processing, among others. In response to the series prediction task, deep learning algorithms automatically discover and extract features from historical data, leading to highly accurate predictions\cite{debroy23, wang23, lam23}. Numerous methods adopt recurrent neural networks (RNNs) and their variants to capture the changing pattern of the data series in the time dimension\cite{chenyt21, kong17, tao19, gao23}. The study \cite{du19} utilizes the strengths of both convolutional networks and long short-term memory (LSTM), and captures complex spatio-temporal relationships in air quality data. The study \cite{liang18} introduces an LSTM-based multi-scale attention network model to selectively focus on different parts of the data at different scales. In the \cite{xiao20, du19}, the weighted ${\rm PM}_{2.5}$ was produced by combining the spatial features with multilayer perception (MLP) or Convolutional Neural Networks (CNN), and historical temporal features were extracted using an LSTM network. 
However, due to the non-grid-based nature of air quality observation stations, the application of convolutional kernels is less effective in extracting information. CNNs are more suitable for processing inputs with regular grid structures, making them less adept at handling the irregular structure of meteorological observation stations.

Consequently, since graph structures are more suitable for describing the distribution and connection of data observation stations, Graph Neural Networks (GNNs) and their variants have been used to capture spatial information interactions recently. The study \cite{qi19} combines graph convolutional neural network (GCN) and LSTM to model the spatial dependencies among monitoring stations and the temporal correlations of historical data to capture the complex dynamics of air pollution. The study \cite{wang20} simulates the ${\rm PM}_{2.5}$ transport between the cities by a knowledge-enhanced GNN and combines Gate Recurrent Unit (GRU) to build a spatio-temporal graph model. The study \cite{gao21} utilizes dynamic graphs with learnable adjacency matrices to detect the spatial correlations at different time points, and LSTM is used to extract the temporal features. These methods improve prediction accuracy by combining RNNs and graph structure, but direct integration is frequently inflexible and poorly thought out. The study \cite{xu21} introduces a city-scale graph on the basis of station-scale graphs, which studies spatial dependence from two scales. 

Although, \cite{xu21} considers the interactions between different spatial scales, the messaging mechanisms are insufficient feature fusion between different scales, and the information interaction between the two scales is not fair. For example, the city-scale graph just only based on the mean of station-scale graphs, which ignores the difference in influence across stations, while the interactions from the city-scale graph to the station-scale graphs are learnable. One scale is information-rich and the other is information-poor, which leads to a larger and larger information gap in the iterative process. Furthermore, and most importantly, to the best of our knowledge, little related work has considered the effect of multi-temporal scale on air quality. In fact, since changes in air quality are always periodic, the information gathered at various time scales is different. For example, a factory always emits polluting gases at 12:00 noon each day, then when predicting the air quality values of a nearby station at 12:00 the following day, we rely more on the data from the previous day's noon and less on the data from the previous moment. Therefore, it is necessary to consider the impact of different time scales on air quality prediction.

In addition, there are also approaches that utilize aerial images \cite{aerial-ganji2020predicting}, satellite data \cite{satellite-xiao2018ensemble}, or street-level images \cite{street-level-feldman2023urban, street-level-xu2022prediction}. However, these methods similarly only consider a single spatial scale. For instance, street-level images typically can only predict air pollution conditions within the urban area, while satellite data often covers larger areas, even spanning multiple countries. It is hard for these types of data sources to consider multiple scales in both spatial and temporal dimensions.

The core challenge in air quality prediction is how to effectively integrate multi-scale information between the station scale and city scale, as well as deal with multi-scale phenomena in the temporal dimension, so as to realize air quality prediction based on spatio-temporal multi-scale information.

In this paper, we propose M2G2, \textbf{M}ulti-spatial \textbf{M}ulti-temporal air quality forecasting method based on \textbf{G}raph Convolutional Networks and \textbf{G}ated Recurrent Units, which considers multi-scale features in both temporal and spatial dimensions and utilizes multiple meteorological indicators to assist in predicting air quality. Specifically, for the spatial dimension, we constructed both a station-scale graph and a city-scale graph, and used bidirectional learnable and residual structures to establish an interaction channel between the two scales, enabling the full integration of features from different scales. For the temporal dimension, we addressed the update of features at different scales in parallel using multiple hidden states. Through dynamic weight assignment, we achieved adaptively integrated temporal features across varying scales. We validate the effectiveness of M2G2 and its components in various aspects through numerous comparative experiments and ablation studies. Additionally, we also demonstrate the effectiveness of M2G2 on three other datasets, besides ${\rm PM}_{2.5}$. 

Our contributions are summarized as follows.
\begin{itemize}
    \item 
    To the best of our knowledge, the proposed air quality prediction method is the first to take into account dual multi-scale in both spatial and temporal dimensions. 
    \item
    We construct a dual-channel learnable multi-spatial scale and dynamic-weight multi-temporal scale network structure M2G2. Based on the station-scale graph and city-scale graph, M2G2 can fuse information between different spatial scales in a bidirectional and learnable way. Using parallel hidden states, M2G2 adaptively fuses information with different temporal scales.
    \item
    We collect air quality data and meteorological data for 41 cities and 152 stations throughout northern China over a five-year period (Jan. 1, 2016 to Aug. 31, 2021). The effectiveness of M2G2 is validated on numerous comparative experiments. The ablation experiments illustrate the effective combination of the various modules of M2G2. We also explore the effect of the choice of different time scales on the results. Furthermore, in long-term prediction, M2G2 exhibits lower relative decay in prediction accuracy compared to other baselines for short-term prediction. Additionally, M2G2 demonstrates excellent performance in predicting different pollutants(${\rm PM}_{2.5}$, ${\rm PM}_{10}$, ${\rm NO}_2$, and ${\rm O}_3$), showcasing its strong generalization ability and the necessity of a multi-scale design.
\end{itemize}

\section{Methodology}

\begin{figure}[htbp]
\centering
\includegraphics[width=0.8\textwidth]{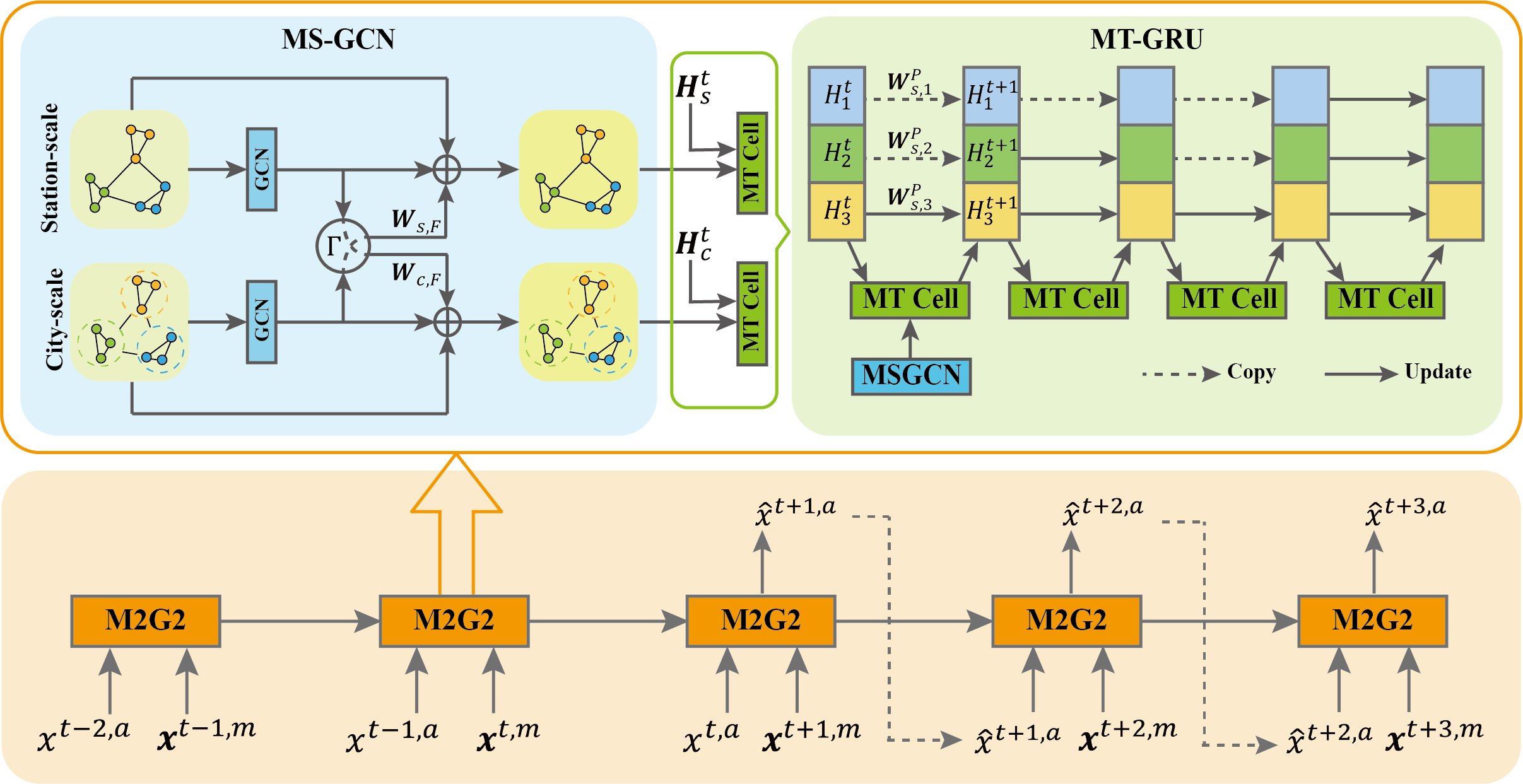}
\caption{An overview of M2G2. \textbf{The orange module}: M2G2 follows a sequence-to-sequence structure, using historical data to predict future air quality. The inputs are air quality and meteorological indicators to predict the air quality at the next moment. M2G2 consists of Multi-scale Spatial GCN (MS-GCN) and Multi-scale Temporal GRU (MT-GRU). In this framework, $t$ means the current time. \textbf{The blue module}: MS-GCN consists of two scales, station and city, and each of the two scales conducts spatial feature extraction by GCN, which results in cross-scale feature interaction. The output of the MS-GCN will be passed into the MT-GRU. \textbf{The green module}: we have improved the update mechanism of the hidden state in GRU by slicing the hidden state and updating it at different intervals. As shown in the figure the hidden state $ \bm{H}^t $ is cut into 3 parts: $ \bm{H}_1^t, \bm{H}_2^t, \bm{H}_3^t $. The solid line represents the update, and the dashed line represents the current iteration step to keep the original value. In addition, $ \bm{H}_1^t, \bm{H}_2^t \ {\rm and} \ \bm{H}_3^t $ each have a learnable dynamic weight, which corresponds to different temporal scales with different significance for the current prediction.}
\label{M2G2}
\end{figure}

As shown in the main part (orange box) of Fig. \ref{M2G2}, we predict the air quality in the future by iteration. At each iteration step, the inputs of M2G2 are the air quality value and meteorological indicator value, and the output is the predicted air quality value for the next iteration step. If the model is forecasting the future, the input air quality value will be replaced by the air quality prediction value from the previous step. As the blue module and green module illustrated in Fig. \ref{M2G2}, the proposed framework consists of two major components: Multi-scale Spatial GCN (MS-GCN) and Multi-scale Temporal GRU (MT-GRU). By modeling different association graphs, MS-GCN captures information at various spatial scales, and allows for interactivity between features across these scales via an assignment matrix following graph convolution. Meanwhile, MT-GRU incorporates multiple hidden states with distinct update periods to capture information at varying temporal scales. In this manner, the M2G2 framework has the ability to adeptly capture and integrate information across multiple spatial and temporal scales concurrently.

\subsection{Problem Definition}

Similarly to previous works \cite{wang20, xiao22}, the utilized dataset encompasses air quality indicators and meteorological indicators, which are attributed to various stations in different cities. Stations and cities can be regarded as nodes, thus forming a graph structure. Additionally, the air quality indicators or meteorological indicators for each station or city form sequences along the temporal dimension. Our objective is to utilize historical data from previous time points to forecast the values of air quality indicators for future time points. Concretely, our problem is defined as follows: Given the assignment matrix $ \bm \Gamma^b $ between cities and stations, the graph of station-scale $ G_s $ and city-scale $ G_c $, the air pollutant concentration $ {\bm X}_s^{1:T, a} \in\mathbb{R}^{T \times S \times 1} $ in station-scale of all the vertices in historical time $T$ and all-time meteorological indicators $ {\bm X}_s^{1:T+\tau, m} \in\mathbb{R}^{(T+\tau) \times S \times M} $ in station-scale of all the vertices. Our goal is to train a function, denoted as $ F(\cdot) $, with the objective of predicting air quality for the next $ \tau $ steps by using historical data.

\vspace{-1.5em}

\begin{align}
    [{\bm X}_s^{1:T, a}, {\bm X}_s^{1:T+\tau, m}; G_s; G_c; {\bm \Gamma^b}] \stackrel{F(\cdot)}{\longrightarrow} {\bm X}_s^{T+1:T+\tau, a}. \nonumber
\end{align} 

\vspace{-1.5em}

\noindent where ${\bm X}_s^{T+1:T+\tau, a}\in\mathbb{R}^{\tau \times S\times 1}$ is the feature air quality values of the station-scale.

Specifically, we represent the $N$ air quality monitoring stations as a graph structure $G_s =\{V_s, E_s, \bm{W}_s\}$, where $V_s$ is the nodes set of stations, $E_s$ is the edges set representing distance among stations, and $S$ is the number of stations. Similarly, we can obtain a graphical representation of the city-scale $G_c=\{V_c, E_c, \bm{W}_c\}$, and $C$ is the number of cities. Each monitoring station observes air quality $ {\bm x}_s^{t, a} \in \mathbb{R}$, where $s$ is the station index, $t$ stands for time, and $a$ represents different air pollutant concentrations, such as ${\rm PM}_{2.5}$, ${\rm PM}_{10}$, ${\rm NO}_2$, ${\rm O}_3$. Analogously, the meteorological indicators are expressed as ${\bm x}_s^{t, m}\in\mathbb{\mathbb{R}}^M$, where $M$ means the number of various meteorological features. The assignment matrix representing the city in which the monitoring station is located is denoted by ${\bm \Gamma^b}$, where ${\bm \Gamma^b}$ is a matrix with shape $S\times C$, $\bm \Gamma^b_{i,j} = 1$ means the $ i^{th} $ station belong to the $j^{th}$ city, $\bm \Gamma^b_{i,j}=0$ means not. The city's air quality ${\bm x}_c^{t,a}$ and meteorological indicators values ${\bm x}_c^{t,m} $ can be aggregated by the assignment matrix where $c$ is the index of cities, and we find that the averaging aggregation is effective in terms of the subsequent experimental results.

\subsection{Graph Construction}

Distance will significantly affect the propagation of air pollutants, the closer two places are, the greater the impact of air quality between them will be, so we calculated the distance matrix $ \bm{W}_s^{dis} $ for station-scale by thresholded Gaussian kernel \cite{shu13}:

\vspace{-1.5em}

\begin{align}
    (\bm{W}_s^{dis})_{ij}:\ = \left\{
    \begin{aligned}
        &exp(-\frac{d_{ij}^2}{\sigma^2})& &,& &\ {\rm for}\ i \neq j\ {\rm and}\ exp(-\frac{d_{ij}^2}{\sigma^2}) \geq \epsilon, & \\
        &0& &,& &\ \rm otherwise. &
    \end{aligned}
    \right. \nonumber
\end{align} 

\vspace{-1.5em}

\noindent where $ d_{ij} $ is the Euclidean distance between $ v_i $ and $ v_j $. $ \sigma^2 $ and $ \epsilon $ are hyperparameters that control distribution and sparsity of $ W^{dis} $. 

\vspace{-1.5em}

\begin{align}
    (\bm{A}_s)_{ij}:\ = \left\{
    \begin{aligned}
        &1& &,& &\ {\rm for} \  (\bm{W}_s^{dis})_{ij} > 0 \ {\rm and} \ {\rm max}_{\gamma\in[0,1]}( h(\gamma \rho_i + (1-\gamma) \rho_j) - h(\rho_i) ) < H, & \\
        &0& &,& &\ \rm otherwise. &
    \end{aligned}
    \right. \nonumber
\end{align} 

\vspace{-1.5em}

\noindent In addition to $ \bm{W}_s^{dis} $, the adjacency matrix $ \bm{A}_s $ is generated by considering the highest elevation between two nodes. If the maximum difference between the altitude of any intermediate point between two nodes and the altitude of the starting node of a directed edge $ e_{ij} $ is less than a threshold $ H $, we consider the two nodes to be linked. The adjacency matrix of the city $ \bm{A}_c $ is calculated by the same method as above.

Regarding the attributes of edges, we take into account the geographical distance between two nodes and their respective orientation.

\subsection{Multi-scale Spatial GCN (MS-GCN)}
\label{msgcn}

According to Fig. \ref{M2G2}, the graph structure has been organized at both the station-scale and city-scale. The graph convolution is used separately on two scales to extract highly meaningful patterns and features in the spatial domain. The computational complexity of graph Fourier-based convolution can reach $ \mathcal{O}(N^2) $, so we use approximation strategies to reduce the expensive overhead. 

~\\
\textbf{${1^{st}}$-order Chebyshev Polynomials Approximation.}\quad The Spectral graph convolution network with graph Fourier transforms is widely applied, which introduces the graph convolution operator $ *_\mathcal{G} $:

\vspace{-1.5em}

\begin{align}
    {\bm L} &= {\bf I}_N - {\bm D}^{-\frac{1}{2}}{\bm A}{\bm D}^{-\frac{1}{2}}, \nonumber\\
    {\Theta} *_\mathcal{G}{\bm x} &= {\Theta}({\bm L}){\bm x} = {\bm U}{\Theta}({\bm\Lambda}){\bm U}^{\top}{\bm x}. \label{originGCN} 
\end{align} 

\vspace{-1.5em}

\noindent where ${\bm L}\in \mathbb{R}^{N \times N} $ is the symmetric normalization of graph Laplacian. $ \bm L $ is calculated by the adjacency matrix ${\bm A}$, identity matrix ${\bf I}_N$ and degree matrix $ \bm D $. The eigenvalue decomposition ${\bm L} = {\bm U}{\bm\Lambda}{\bm U}^{\top}$ of $ \bm L $ yields the eigenvector matrix $ \bm U \in \mathbb{R}^{N \times N} $, which serves as the basis of graph Fourier transform. $ \bm\Lambda \in \mathbb{R}^{N \times N} $ is the diagonal matrix consisting of eigenvalues. The inputs of graph convolution operator $ *_\mathcal{G} $ are a signal $ x \in \mathbb{R}^N $ and a learnable convolution kernel $ \Theta $. 

Since Eq. (\ref{originGCN}) will introduce a high computational complexity, huge parameters and global receptive field, using Chebyshev polynomials for fitting convolution kernels is widely used that can minimize the complexity and localize the filter's field. The graph convolution kernel will be approximated by Chebyshev polynomials as follows:

\vspace{-1.5em}

\begin{align}
    \Theta(\bm\Lambda) \approx \sum_{k=0}^{K-1} \theta_k T_k(\Tilde{\bm\Lambda}). \nonumber
\end{align}

\vspace{-1.5em}

\noindent where $ \theta_k $ is the learnable coefficient that needs to be iteratively updated. $ T_k(\cdot) $ is the Chebyshev polynomial of order k. $ \Tilde{\bm\Lambda} = 2\bm\Lambda / \lambda_{max} - {\bf I}_N $ rescales $ \bm \Lambda $ to ensure that the input of the Chebyshev polynomial in $ [-1, 1] $ by maximum eigenvalue $ \lambda_{max} $. 

The Eq. (\ref{originGCN}) can be rewritten as:

\vspace{-1.5em}

\begin{align}
    \Theta *_\mathcal{G} x & \approx \bm U \sum_{k=0}^{K-1} \theta_k T_k(\Tilde{\bm \Lambda})) \bm {U}^{\top}x
     = \sum_{k=0}^{K-1} \theta_k T_k(\bm U \Tilde{\bm \Lambda} \bm{U}^{\top}) x 
     = \sum_{k=0}^{K-1} \theta_k T_k(\Tilde{\bm L}) x. \label{chebGCN} 
\end{align}

\vspace{-1.5em}

\noindent where $ \Tilde{\bm L} = 2\bm L / \lambda_{max} - {\bf I} $. For Eq. (\ref{chebGCN}), the graph Laplacian matrix $ L $ does not require eigenvalue decomposition. The cost will be reduced to $ \mathcal{O}(K|E|) $. In addition, the Chebyshev polynomial has the following property:

\vspace{-1.5em}

\begin{gather}
    T_k(\Tilde{\bm L}) = 2 \Tilde{\bm L} T_{k-1}(\Tilde{\bm L}) - T_{k-2}(\Tilde{\bm L}), \nonumber \\
    T_0(\Tilde{\bm L}) = {\bf I}_N, \ T_1(\Tilde{\bm L}) = \Tilde{\bm L}. \nonumber
\end{gather}

\vspace{-1.5em}

\noindent $ K $-localized convolutions will aggregate information about the $ (K-1) $-order neighbors of the object node. The $ 1^{st} $-order aggregation operation is cost-effective on large-scale graphs, and stacking $ 1^{st} $-order aggregations can expand the neighborhood of the graph convolution. Furthermore, we can assume that $ \lambda_{max} = 2 $, the ensuing scaling effect can be automatically adapted through network learning. When $ K = 2 $ ($ 1^{st} $-order aggregation), Eq. (\ref{chebGCN}) can be overwritten as:

\vspace{-2.em}

\begin{align}
    \Theta *_\mathcal{G} x &\approx \theta_0 x + \theta_1 \Tilde{\bm L} x 
    \approx \theta_0 x + \theta_1 (\bm L - {\bf I}_N) x 
    = \theta_0 x - \theta_1 \bm D^{-\frac{1}{2}}\bm A \bm D^{-\frac{1}{2}} x. \nonumber
\end{align}

\noindent Next, let $ \theta = \theta_0 = -\theta_1$ to enhance numerical stability. To further alleviate numerical instabilities
and exploding/vanishing gradients, the renormalization trick is introduced: $ \bm A $ transforms to $ \Tilde{\bm A} = \bm A + {\bf I}_N $, the corresponding degree matrix $ \Tilde{\bm D} $ will be recalculated with $ \Tilde{\bm D}_{ii} = \sum_j \bm A_{ij} $. Finally, the graph convolution operator can be expressed:

\begin{align}
    \Theta *_\mathcal{G} x &= \theta ( {\bf I}_N + \bm D^{-\frac{1}{2}}\bm A \bm D^{-\frac{1}{2}} ) x
    = \theta ( \Tilde{\bm D}^{-\frac{1}{2}} \Tilde{\bm A} \Tilde{\bm D}^{-\frac{1}{2}} ) x. \nonumber
\end{align}

\noindent The above definition can be extended to $ \bm X \in \mathbb{R}^{N \times C} $ with $ C_{in} $ input channels and $ C_{out} $ output channels as follows:

\begin{align}
    Y = \sigma ( \Tilde{\bm D}^{-\frac{1}{2}} \Tilde{\bm A} \Tilde{\bm D}^{-\frac{1}{2}} \bm X \bm W). \label{commonGCN} 
\end{align}

\noindent where $ \bm W \in \mathbb{R}^{C_{in} \times C_{out}} $ is the learnable convolutional kernel matrix and $ \bm Y \in \mathbb{R}^{N \times C_{out}} $ is the output of graph convolution operator after an activation function $ \sigma (\cdot) $. 

\vspace{1em}
\begin{table}[htbp]
\centering
\begin{tabular}{p{16cm}}
\toprule
{\bf Algorithm 1:} Multi-scale Spatial GCN (MS-GCN)\\
\midrule
\hangafter 1
\hangindent 1.75em
\enspace 1: {\bf Input:} Batch size $b$, Monitoring stations features $ \bm{X}_s^t $, Cities' features $ \bm{X}_c^t $, Assignment matrix $ \bm \Gamma^b $, Learnable GCN weight matrices of station-scale and city-scale $ \bm{W}_{s,GCN}, \bm{W}_{c,GCN} $, Learnable cross-scale transform matrices $ \bm{W}_{s,F},  \bm{W}_{c,F} $ \\
\enspace 2: {\bf while} $ \bm{W}_{s,GCN}, \bm{W}_{c,GCN}, \bm{W}_{s,GCN}, \bm{W}_{c,GCN} $ not converged {\bf do} \\
\enspace 3: \quad Sample $ \bm{X}_s^t , \bm{X}_c^t $ from the training data with $b$ instances \\
\enspace 4: \quad // Graph Convolution on Two Scales\\
\hangafter 1
\hangindent 3.75em
\enspace 5: \qquad Generate $ \bm{X}_{s,GCN}^t $ by applying graph convolution on the station-scale features $ \bm{X}_s^t $ according to Eq. (\ref{commonGCN}) through $ \bm{W}_{s,GCN} $. \\
\hangafter 1
\hangindent 3.75em
\enspace 6: \qquad Generate $ \bm{X}_{c,GCN}^t $ by applying graph convolution on the city-scale features $ \bm{X}_c^t $ according to Eq. (\ref{commonGCN}) through $ \bm{W}_{c,GCN} $. \\
\enspace 7: \quad // Station-City Bidirectional-Fusion Module \\
\hangafter 1
\hangindent 3.75em
\enspace 8: \qquad $\bm{X}_{s,F}^t = [\bm{X}_s^t, \ \bm{\Gamma}^b \bm{X}_{c,GCN}^t \bm{W}_{s,F}]$ \quad $\triangleright$ Transfer the GCN features $ \bm{X}_{c,GCN}^t $ at city-scale to station-scale by utilizing assignment matrix $ \bm \Gamma^b $ and learnable matrix $ \bm{W}_{s,F} $. \\
\hangafter 1
\hangindent 3.75em
\enspace 9: \qquad $\bm{X}_{c,F}^t = [\bm{X}_c^t, \ {\bm{\Gamma}^b}^{\top} \bm{X}_{s,GCN}^t \bm{W}_{c,F}]$ \quad $\triangleright$ Transfer the GCN features $ \bm{X}_{s,GCN}^t $ at station-scale to city-scale by utilizing assignment matrix $ {\bm{\Gamma}^b}^{\top} $ and learnable matrix $ \bm{W}_{c,F} $. \\
\hangafter 1
\hangindent 2.75em
10: \quad Pass the features learned by MS-GCN into the subsequent MT-GRU as depicted in Sec. (\ref{mtgru}), and then obtain the final prediction results. \\
\hangafter 1
\hangindent 2.75em
11: \quad Compute the loss described in Sec. (\ref{loss}) and update all learnable weights by backpropagating gradients. \\
12: {\bf end while} \\
\hangafter 1
\hangindent 1.75em
13: Calculate the MAE and RMSE using the above prediction results and the ground truth. \\
14: {\bf return} the final learned model. \\

\bottomrule
\label{a1}
\end{tabular}
\end{table}

\textbf{Graph Convolutions and Information Interaction on Two Scales.}\quad As shown in the blue module (MS-GCN) of Fig. \ref{M2G2}, the city-scale input and the station-scale input will be fed into the graph convolution layer separately to learn the spatial features. The two scales of graph convolution will lead to feature extraction at different spatial granularities. The station-scale convolution operation effectively captures the flow of pollutants within the same city, while the city-scale convolution tends to reflect the interactions of air quality between cities, which facilitates us to grasp the global features and local features to make more accurate predictions. Referring to Eq. (\ref{commonGCN}), the two different scales' feature extraction can be described as:

\begin{align}
    & \bm{X}_{s,GCN}^t = \sigma ( \Tilde{\bm{D}}_s^{-\frac{1}{2}} \Tilde{\bm{A}}_s \Tilde{\bm{D}}_s^{-\frac{1}{2}} \bm{X}_s^t \bm{W}_{s,GCN}), \nonumber \\
    & \bm{X}_{c,GCN}^t = \sigma ( \Tilde{\bm{D}}_c^{-\frac{1}{2}} \Tilde{\bm{A}}_c \Tilde{\bm{D}}_c^{-\frac{1}{2}} \bm{X}_c^t \bm{W}_{c,GCN}). \nonumber
\end{align}

\noindent where $ \bm{X}_s^t \in \mathbb{R} ^ {S \times (M + 1)}$ denotes all monitoring stations features, obtained from the previous air quality $ \bm{X}_s^{t-1, a} \in\mathbb{R}^{S\times 1} $ and current meteorological indicators $ \bm{X}_s^{t, m} \in\mathbb{R}^{S\times M} $, $ \bm{X}_c^t \in \mathbb{R} ^ {C \times (M + 1) } $ is all cities' features aggregated by the assignment matrix $ \bm \Gamma^b $. $ \bm{X}_{s,GCN}^t \in \mathbb{R} ^ {S \times C_{out}^{GCN}}, \ \bm{X}_{c,GCN}^t \in \mathbb{R} ^ {C \times C_{out}^{GCN} } $ is the graph convolution result of two scales by the trainable weight matrix $ \bm{W}_{s,GCN}, \bm{W}_{c,GCN} \in \mathbb{R} ^ {(M + 1) \times C_{out}^{GCN} } $.

For the station-scale, the perception range can be expanded by introducing city-scale features. In turn for city-scale, station characteristics are significant for an accurate prediction of current city's air quality. So we propose the Station-City Bidirectional-Fusion Module to complete information interaction between the two scales:

\begin{gather}
    \bm{X}_{s,F}^t = [\bm{X}_s^t, \ \bm{\Gamma}^b \bm{X}_{c,GCN}^t \bm{W}_{s,F}], \nonumber \\
    \bm{X}_{c,F}^t = [\bm{X}_c^t, \ {\bm{\Gamma}^b}^{\top} \bm{X}_{s,GCN}^t \bm{W}_{c,F}]. \label{s2cFusion} 
\end{gather}

\noindent where $ [\cdot , \cdot] $ is concatenation. Given the assignment matrix $ \bm{\Gamma}^b $ , the learnable transform matrixes $ \bm{W}_{s,F},  \bm{W}_{c,F} \in \mathbb{R} ^ { C_{out}^{GCN} \times C_{out}^{F} }$ achieve cross-scale transfer of spatial features. The final outputs of Station-City Bidirectional-Fusion Module $ \bm{X}_{s,F}^t \in \mathbb{R} ^ { S \times {(C_{out}^{F}+M+1)} },  \bm{X}_{c,F}^t \in \mathbb{R} ^ { C \times {(C_{out}^{F}+M+1)} }$ concatenates the origin input. The pseudocode of MS-GCN module is shown in Algorithm 1.

\subsection{Multi-scale Temporal GRU (MT-GRU)}
\label{mtgru}

GRU is a widely used recurrent neural network based gate, which has a specialized learnable mechanism to determine when the hidden state should be updated, and when the hidden state should be reset. This mechanism is used to solve the long-term memory problem. However, time series data frequently exhibit distinct temporal scale properties that were not intended to be taken into account by the original GRU. As shown in the green module (MT-GRU) of Fig. \ref{M2G2}, we propose a variant of GRU that modifies the update mechanism of the hidden state. The update intervals of different parts of the hidden state are inconsistent, so that features at different temporal scales can be extracted explicitly. The detailed process will be further described in subsequent paragraphs using the station-scale as an example, which is consistent with the city-scale.

We divide the hidden state into $ V $ different temporal update scales. Since the importance of each scale should change for the prediction of the present instant, we will first calculate the dynamic temporal scale weights $ \bm{W}_s^P \in \mathbb{R}^V $. Each temporal scale has a corresponding weight.

\begin{gather}
    \bm{W}_s^P = \sigma ( \bm{X}_{s,F}^t \bm{W}_{xp} + \bm{H}_s^{t-1} \bm{W}_{hp} + \bm{b}_p ). \label{dynamicW} 
\end{gather}

We divide the hidden state $ \bm{H}_s^{t-1} \in \mathbb{R}^{S \times C_h} $ equally into $ V $ parts by channel dimension ($ 0 \equiv C_h \bmod{V} $). $ \bm{H}_s^{t-1} $ can be expressed as $ \bm{H}_s^{t-1} = [\bm{H}_{s,1}^{t-1}, \bm{H}_{s,2}^{t-1}, \cdots, \bm{H}_{s,V}^{t-1}] $. We update $ \bm{H}_s^{t-1} $ with the obtained weights $ \bm{W}_s^P $, the new weighted hidden state matrix $ \bm{H}_s^{' t-1} = [\bm{W}_{s,1}^P \bm{H}_{s,1}^{t-1}, \bm{W}_{s,2}^P \bm{H}_{s,2}^{t-1}, \cdots, \bm{W}_{s,V}^P \bm{H}_{s,V}^{t-1}] $


After that we can calculate the reset gate $ \bm{R}_s^t \in \mathbb{R}^{S \times C_h} $, update gate $ \bm{Z}_s^t \in \mathbb{R}^{S \times C_h} $ and the candidate hidden state $ \Tilde{\bm{H}}_s^t \in \mathbb{R}^{S \times C_h} $ following the original GRU:

\vspace{-2.5em}

\begin{gather}
    \bm{R}_s^t = \sigma ( \bm{X}_{s,F}^t \bm{W}_{xr} + \bm{H}_s^{' t-1} \bm{W}_{hr} + \bm{b}_r ), \nonumber \\
    \bm{Z}_s^t = \sigma ( \bm{X}_{s,F}^t \bm{W}_{zr} + \bm{H}_s^{' t-1} \bm{W}_{hz} + \bm{b}_z ), \nonumber \\
    \Tilde{\bm{H}}_s^t = {\rm tanh} ( \bm{X}_{s,F}^t \bm{W}_{xh} + (\bm{R}_s^t \odot \bm{H}_s^{' t-1}) \bm{W}_{hh} + \bm{b}_h ). \nonumber
\end{gather}

\vspace{-1em}

Similarly to the hidden state division, we divide the new weighted hidden state matrix $ \bm{H}_s^{' t-1} \in \mathbb{R}^{S \times C_h} $, the candidate hidden state $ \Tilde{\bm{H}}_s^t $ and the update gate $ \bm{Z}_s^t $ equally into $ V $ parts by channel dimension:

\vspace{-2.5em}

\begin{gather}
    \bm{H}_s^{' t-1} = [\bm{H}_{s,1}^{' t-1}, \cdots, \bm{H}_{s,V}^{' t-1}], \nonumber \\
    \Tilde{\bm{H}}_s^t = [\Tilde{\bm{H}}_{s,1}^t, \cdots, \Tilde{\bm{H}}_{s,V}^t], \nonumber \\
    \bm{Z}_s^t = [\bm{Z}_{s,1}^t, \cdots, \bm{Z}_{s,V}^t]. \nonumber
\end{gather}

\vspace{-1em}

\noindent where $ \bm{H}_{s,v}^{' t-1}, \Tilde{\bm{H}}_{s,v}^t, \bm{Z}_{s,v}^t $ respectively means taking slices from $ ((v-1) \times C_h / V) $ to $ (v \times C_h / V) $ along the feature channel of $ \bm{H}_s^{' t-1}, \Tilde{\bm{H}}_s^t, \bm{Z}_s^t $. $ v $ represents the index of $V$ parts, which belongs to a range from $ 1 $ to $ V $.

Next we define the temporal scale vector $ \bm{P} \in \mathbb{N}_1^{V} = [P_1, P_2, \cdots, P_V] $. $ P_v $ represents the $ v^{th} $ part's update periods. The original GRU hidden state update mechanism will be rewritten as:

\vspace{-2em}
\begin{align}
    \bm{H}_{s,v}^t:\ = \left\{
    \begin{aligned}
        &\bm{Z}_{s,v}^t \odot \bm{H}_{s,v}^{' t-1} + (1 - \bm{Z}_{s,v}^t) \odot \Tilde{\bm{H}}_{s,v}^t&,\ & \ {\rm for} \ t \bmod{P_v} =0, \nonumber \\
        &\bm{H}_{s,v}^{' t-1}&,\ & \ {\rm otherwise}. \nonumber
    \end{aligned}
    \right.
\end{align}

\vspace{-2em}

\begin{table}[H]
\centering
\begin{tabular}{p{16.cm}}
\toprule
{\bf Algorithm 2:} Multi-scale Temporal GRU (MT-GRU)\\
\midrule
\hangafter 1
\hangindent 1.75em
\enspace 1: {\bf Input:} Stations features from MS-GCN module $ \bm{X}_{s,F}^t $, Learnable weight matrices $ \bm{W}_{xp}, \bm{W}_{hp}, \bm{b}_p $, Dynamic temporal scale weights $ \bm{W}_s^P $, Previous step's hidden state $ \bm{H}_s^{t-1} $, New weighted hidden state $ \bm{H}_s^{' t-1} $, Reset gate $ \bm{R}_s^t $, Update gate $ \bm{Z}_s^t $, Candidate hidden state $ \Tilde{\bm{H}}_s^t $, Temporal scale vector $ \bm{P} $, Number of parts $ V $, Current step's hidden state $ \bm{H}_s^{t} $\\
\enspace 2: {\bf while} $ \bm{W}_{xp}, \bm{W}_{hp}, \bm{b}_p $ and other learnable weights of GRU not converged {\bf do} \\
\enspace 3: \quad Get $ \bm{X}_{s,F}^t $ from the MS-GCN module depicted in Sec. (\ref{msgcn}) \\
\hangafter 1
\hangindent 2.75em
\enspace 4: \quad $\bm{W}_s^P = \sigma ( \bm{X}_{s,F}^t \bm{W}_{xp} + \bm{H}_s^{t-1} \bm{W}_{hp} + \bm{b}_p )$ \quad $\triangleright$ Calculate the dynamic temporal scale weights $ \bm{W}_s^P $ by feed-forward network ($ \bm{W}_{xp}, \bm{W}_{hp}, \bm{b}_p $) and previous step's hidden state $ \bm{H}_s^{t-1} $. \\
\hangafter 1
\hangindent 2.75em
\enspace 5: \quad $ \bm{H}_s^{' t-1} = [\bm{W}_{s,1}^P \bm{H}_{s,1}^{t-1}, \bm{W}_{s,2}^P \bm{H}_{s,2}^{t-1}, \cdots, \bm{W}_{s,V}^P \bm{H}_{s,V}^{t-1}] $ \quad $\triangleright$ Divide previous step's hidden state $ \bm{H}_s^{t-1} $ into $ V $ parts and scale each part using the corresponding dynamic weights $ \bm{W}_s^P $ to obtain new weighted hidden state $ \bm{H}_s^{' t-1} $. \\
\hangafter 1
\hangindent 2.75em
\enspace 6: \quad Calculate reset gate $ \bm{R}_s^t $, update gate $ \bm{Z}_s^t $ and candidate hidden state $ \Tilde{\bm{H}}_s^t $ in the manner of original GRU. \\
\enspace 7: \quad {\bf if} ${\rm for} \ t \bmod{P_v} =0$ {\bf then} \\
\hangafter 1
\hangindent 3.75em
\enspace 8: \qquad $\bm{H}_{s,v}^t:\ = \bm{Z}_{s,v}^t \odot \bm{H}_{s,v}^{' t-1} + (1 - \bm{Z}_{s,v}^t) \odot \Tilde{\bm{H}}_{s,v}^t$ \quad $\triangleright$ Follow defined temporal scale vector $ \bm{P} \in \mathbb{N}_1^{V} = [P_1, P_2, \cdots, P_V] $. If $ t \bmod{P_v} =0, \nonumber $ i.e. the $ v $-th part $ \bm{H}_{s,v}^t $ is determined to be updated via reset gate $ \bm{R}_s^t $, update gate $ \bm{Z}_s^t $ and candidate hidden state $ \Tilde{\bm{H}}_s^t $ at the current time step $ t $. \\
\enspace 9: \quad {\bf else} \\
\hangafter 1
\hangindent 3.75em
10: \qquad $\bm{H}_{s,v}^t:\ = \bm{H}_{s,v}^{' t-1}$ \quad $\triangleright$ Otherwise, $ \bm{H}_{s,v}^t $ retains its original value.\\
11: \quad {\bf end if} \\
\hangafter 1
\hangindent 2.75em
12: \quad Compute the loss described in Sec. (\ref{loss}) and update all learnable weights by backpropagating gradients. \\
13: {\bf end while} \\
\hangafter 1
\hangindent 1.75em
14: Calculate the MAE and RMSE using the above prediction results and the ground truth. \\
15: {\bf return} the final learned model. \\

\bottomrule
\label{a2}
\end{tabular}
\end{table}

\vspace{-2em}

Due to the independence of the MT-GRU modules at the station-scale and city-scale, the pseudocode for the optimization process shown in Algorithm 2 only presents the procedure at the station-scale. The pseudocode for the city-scale is consistent.

\vspace{-1.em}

\subsection{Air quality concentration prediction and objective function}
\label{loss}

Following multi-scale spatial and multi-scale temporal modules, we use a single-layer feed-forward network to predict air quality:

\vspace{-2.em}
\begin{align}
    \hat{\bm{X}}_s^{t, a} = \sigma ( \bm{H}_{s,v}^t \bm{W}_{s,ha} + \bm{b}_{s,a} ). \nonumber \\
    \hat{\bm{X}}_c^{t, a} = \sigma ( \bm{H}_{c,v}^t \bm{W}_{c,ha} + \bm{b}_{c,a} ). \nonumber
\end{align}
\vspace{-1.em}

We use the MSE loss for the prediction task:

\vspace{-2.em}
\begin{align}
    loss = \frac{1}{S} \frac{1}{\tau} \sum_{t=T+1}^{T + \tau} {\Vert \bm{X}_s^{t, a} -  \hat{\bm{X}}_s^{t, a} \Vert}_2^2 + \frac{1}{C} \frac{1}{\tau} \sum_{t=T+1}^{T + \tau} {\Vert \bm{X}_c^{t, a} -  \hat{\bm{X}}_c^{t, a} \Vert}_2^2. \nonumber
\end{align}

\vspace{-1.em}
\noindent where $loss$ is utilized to train the MS-GCN and MT-GRU through gradient backpropagation.

\vspace{-1.em}

\subsection{Intuitive understanding of M2G2}

\vspace{-1.em}

\begin{figure}[H]
\centering
\includegraphics[width=0.9\textwidth]{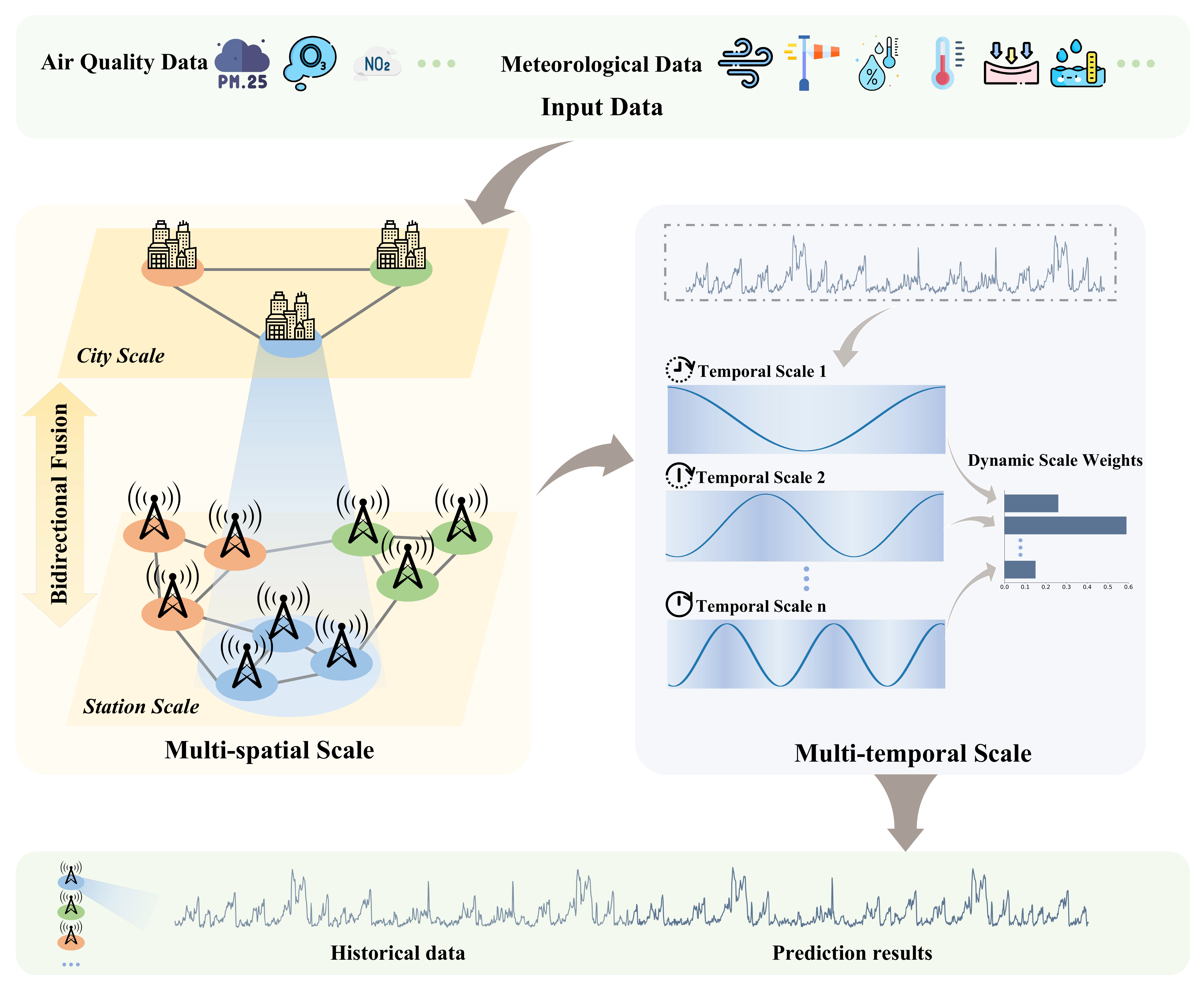}
\caption{The core idea of M2G2: At the spatial scale, we employ the bidirectional fusion module to learn feature information that mutually enhances the city-scale and station-scale, using an end-to-end approach. After spatial feature extraction, the relevant information is fed into the temporal dimension module. In this module, components of different scales are utilized to extract time-series features with distinct periodicities. Finally, these features are aggregated using dynamically learnable weights and produce the final predictions.}
\label{2.6}
\end{figure}

As shown in Fig. \ref{2.6}, M2G2 performs spatial feature learning at both the station and city scales, followed by information interaction through bidirectional fusion module, which effectively utilizes feature representations from different spatial scales; In the temporal dimension, the time series of pollutant concentrations exhibit various periodic scales. To capture this phenomenon, we designed different update frequencies, while assigning dynamic and learnable importance to different scales.

Specifically, for spatial feature extraction, due to the irregular distribution of air quality monitoring stations within a region, the sampled data is sparse and unevenly distributed. GCN can effectively handle such irregular data by operating on graph structures. The edges in the graph represent spatial relationships between different monitoring stations, such as distance, wind direction, and other geographical features. GCN can actively learn to extract meaningful representations while aggregating information, thereby capturing complex relationships that may not be apparent in the raw data. We not only consider the spatial dependencies at the station scale but also believe that there exist spatial interactions between cities. Therefore, we introduce a multi-scale structure based on GCN. When considering the interaction between station-scale and city-scale spatial features, previous methods have relied solely on non-learnable one-way mappings through assignment matrices\cite{xu21}. In order to fully leverage the features from two different spatial scales, we propose the design of bidirectional learnable channels that can maximally fit the data distribution and better learn the underlying relationships between different spatial scales. 

Currently, in time series algorithms, GRU offers advantages such as a smaller number of parameters, a simpler structure, and the ability to alleviate the vanishing gradient problem. However, existing research\cite{qi19, wang20, gao21, xu21} has overlooked the fact that the temporal dimension also exhibits a multi-scale phenomenon. We improve the update mechanism of GRU to explore different periodic scales for time series data. Specifically, the hidden state of GRU is decomposed into parts with different update frequencies. Additionally, different periodic scales are associated with adaptive weights, allowing the network to autonomously adjust the distribution of importance across different time scales at different time points.


\section{Experiments}
In this section, we conduct extensive experiments on real-world data to demonstrate the effectiveness of M2G2. Additionally, we provide comprehensive implementation details and analysis based on experimental results.

\subsection{Experimental Setting}

\subsubsection{Dataset Description}
\label{section: dataset}

\begin{figure}[H]
\centering
\includegraphics[width=0.7\textwidth]{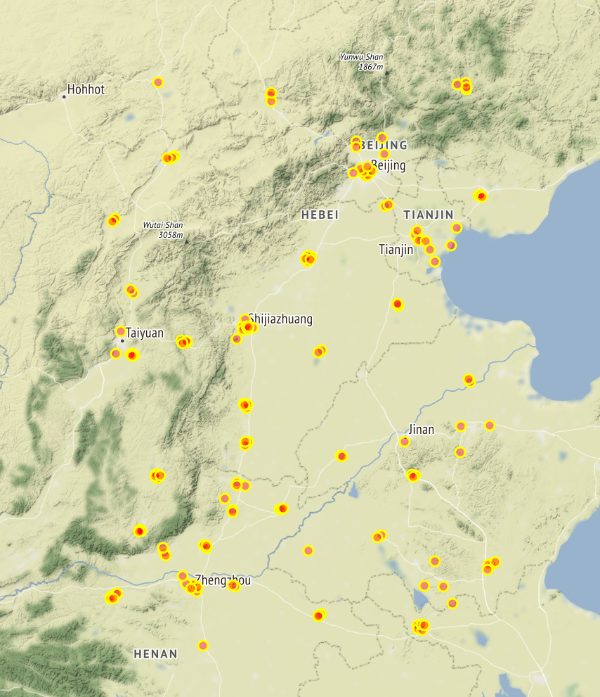}
\caption{The map of the distribution of air quality monitoring stations}
\label{stationMap}
\end{figure}

We collected air quality data and meteorological data for 41 cities throughout northern China over a five-year period (Jan. 1, 2016 to Aug. 31, 2021). Our study focuses on a geographic area primarily centered around Beijing, which includes several key cities and is known for its high scales of air pollution. This region is home to a network of 152 air quality monitoring stations, which are distributed across multiple areas of China and are depicted in Fig. \ref{stationMap}. These stations provide a wealth of data that can be leveraged to gain insights into the spatial and temporal patterns of air pollution in the region.

\begin{itemize}
    \item \textit{Air quality data}: Each of the 152 air quality monitoring stations in our study area collects hourly measurements of four key pollutants: ${\rm PM}_{2.5}$, ${\rm PM}_{10}$, ${\rm NO}_2$, and ${\rm O}_3$, which are obtained from ministry of ecology and environment (MEE)\footnote{\url{https://english.mee.gov.cn/}}. These pollutants are known to have significant impacts on biodiversity, particularly in large cities and areas close to industrial sources. In contrast to previous studies, which have typically focused on one pollutant, we included all four contaminants as prediction targets in our experiments to demonstrate the effectiveness of our proposed method.
    \item \textit{Meteorological data}: The meteorological data collected from ERA5 \footnote{\url{https://cds.climate.copernicus.eu/cdsapp\#!/dataset/reanalysis-era5-single-scales}}, which is the climate reanalysis produced by European Centre for Medium-Range Weather Forecasts (ECMWF), providing boundary layer height, surface pressure, temperature, relative humidity, precipitation, wind speed, wind direction and dew point temperature.
\end{itemize}

A large proportion of monitoring stations have more serious missing data, the missing data rate of the 152 stations we screened was less than 15\%. To handle the missing data, we employed a K-Nearest Neighbor (KNN) interpolation approach based on spatio-temporal similarity. Following prior research, we estimated the concentration of air pollutants for the next 24 steps based on the readings from the previous 24 steps, with each step representing a 3-hour interval. In other words, we made a 72-hour projection based on data from the previous 72 hours. To facilitate model convergence and improve stability, we normalized all model inputs using the Z-score normalization method. We split the entire dataset into three subsets for training, validation, and testing, respectively, with the time periods being (2016/09/01 to 2019/08/31), (2019/09/01 to 2020/08/31), and (2020/09/01 to 2021/08/31).

\subsubsection{Implementation Details}

All experiments are performed on a Slurm cluster with 8 NVIDIA V100 32GB GPUs. Our model and all the baselines are implemented with PyTorch 1.13.1 and pytorch\_geometric (PyG) 2.2.0. To ensure a fair comparison between models, we keep all common settings constant and run each method five times with varying seed values ranging from 1 to 5. By computing the mean value of the results from these multiple runs, we are able to obtain a more reliable estimate of the model performance, while also minimizing the impact of random fluctuations that may occur during the training process. The Adam optimizer is utilized to train models with a learning rate $lr=1e^{-4}$. We train models for 50 epochs with batch size of 64, and early stopping is also adopted on the validation loss. For the temporal scale vector $ {\bm{P}} $ in MT-GRU, we conduct a grid search and $ [1, 2, 4] $ is the best. The mean square error (MSE) between the estimator and the ground truth is employed as the loss function and minimized using backpropagation.

\subsubsection{Evaluation Metrics}

Referring to previous work \cite{wang20, xiao22}, We use Root Mean Square Error (RMSE) and Mean Absolute Error (MAE) defined as follows as the evaluation metrics.

\begin{gather}
    {\rm RMSE} =  \sqrt { \frac{1}{S} \frac{1}{\tau} \sum_{s=1}^{S} {\sum_{t=T + 1}^{T + \tau} (  x_s^{t, a} -  \hat{x}_s^{t, a} )^2 }}, \nonumber \\
    {\rm MAE} =  \frac{1}{S} \frac{1}{\tau} \sum_{s=1}^{S} {\sum_{t=T + 1}^{T + \tau} { | x_s^{t, a} -  \hat{x}_s^{t, a} | }}. \nonumber
\end{gather}

\subsubsection{Baselines for Comparison}

\begin{itemize}
    \item \textbf{GC-GRU\cite{li18}}: Graph Convolutional Gated Recurrent Unit (GC-GRU) is a variant of the Gated Recurrent Unit (GRU) that is designed to work on graph-structured data. GC-GRU combines the GRU architecture with graph convolutional neural networks (GCNs), which allow for information propagation between nodes in a graph. The sizes of the GRU hidden state and the GCN output dimension are 32 and 1 respectively.
    \item \textbf{GC-LSTM\cite{qi19}}: Similar to GC-GRU, Graph Convolutional Long Short-Term Memory (GC-LSTM) is a neural network architecture that combines the concepts of graph convolutional networks (GCN) and long short-term memory (LSTM) networks to operate on graph-structured data. The sizes of the hidden state and the output dimension are 32 and 1 respectively.

    \item \textbf{Graph WaveNet\cite{wu19}}: Graph WaveNet employs dilated convolution to acquire temporal dependencies and trains a new adjacency matrix depending on the data to acquire spatial information. In the Graph WaveNet, the dimensions of the residual channel, dilation channel, skip channel and end channels are 32, 32, 256 and 512 separately. In addition, the number of stacked layers of spatio-temporal convolution is set to 4.

    \item \textbf{GAGNN\cite{chen21}}: The group-aware graph neural network (GAGNN) learns correlations between city groups to effectively capture dependencies between city groups. The hidden size of GNN is set to 32 and the layer number of GNN is set to 2.
    \item \textbf{STGCN\cite{yu18}}: In contrast to the prior approach, the spatio-temporal graph convolutional network (STGCN) employs CNN rather than the widely utilized RNN structure in the temporal feature dimension. The STGCN consists of two spatio-temporal convolutional blocks (ST-Conv blocks). In the ST-Conv block, the dimensions of temporal gated convolution layers and spatial graph convolution layer are set to 64 and 16 respectively.
    \item \textbf{ASTGCN \& MSTGCN\cite{guo19}}: A spatio-temporal attention module that can dynamically describe spatial and temporal relationships is implemented by the attention based spatial-temporal graph convolutional network (ASTGCN). In addition, the recent segment, the daily-periodic segment, and the weekly-periodic segment are three temporal characteristic modules that are produced. It becomes the multi-component spatial-temporal graph convolution network (MSTGCN) when the spatio-temporal attention module is removed. The hyperparameter settings of ASTGCN and MSTGCN are the same to those of STGCN.
    \item \textbf{$ \bf PM_{2.5}\text{-}GNN $\cite{wang20}}: The $ \rm PM_{2.5}\text{-}GNN $ introduce domain knowledge for explicit long-term modeling, which uses wind speed, wind direction and relative position to calculate the advection coefficient. The sizes of the hidden state and the output dimension are 32 and 1 respectively.
    \item \textbf{HighAir\cite{xu21}}: To facilitate efficient information sharing between stations in various cities, HighAir incorporates historical information about the air quality of nearby cities into the station-scale map. Nevertheless, it is unable to properly utilize spatial multi-scale information because it lacks effective learnable components. The hidden size of GNN is set to 32, and the hidden state size of LSTM is set to 64.
\end{itemize}

To ensure a fair comparison, we tune different hyperparameters for each baseline, determining the optimal setting for each.

\subsection{Comparison Study}

\begin{table}[htbp]
\caption{${\rm PM}_{2.5}$ baseline table. 1-24h, 25-48h, and 49-72h represent the performance of predicting pollutant concentrations for the next 1-24 hours, 25-48 hours, and 49-72 hours, respectively.}
\centering
\resizebox{\textwidth}{!}{ 
\begin{tabular}{ccccccc}
\toprule
\multirow{2}*{Model} & \multicolumn{2}{c}{\textbf{\underline{\qquad\quad 1-24h\quad\qquad}}} & \multicolumn{2}{c}{\textbf{\underline{\quad\qquad 25-48h\qquad\quad}}} & \multicolumn{2}{c}{\textbf{\underline{\quad\qquad 49-72h\qquad\quad}}} \\
&MAE&RMSE&MAE&RMSE&MAE&RMSE\\
\midrule
GCGRU\cite{li18} & 16.65 $\pm$ 0.33 & 19.43 $\pm$ 0.35 & 19.97 $\pm$ 0.22 & 22.78 $\pm$ 0.25 & 21.35 $\pm$ 0.25 & 24.15 $\pm$ 0.28\\
STGCN\cite{yu18} & 21.97 $\pm$ 1.39 & 24.83 $\pm$ 1.37 & 25.02 $\pm$ 0.81 & 27.82 $\pm$ 0.79 & 26.91 $\pm$ 0.81 & 29.69 $\pm$ 0.79\\
GWNET\cite{wu19} & 23.89 $\pm$ 0.33 & 26.72 $\pm$ 0.34 & 25.68 $\pm$ 0.38 & 28.73 $\pm$ 0.36 & 26.64 $\pm$ 0.28 & 29.39 $\pm$ 0.29\\
GCLSTM\cite{qi19} & 17.26 $\pm$ 0.87 & 20.04 $\pm$ 0.89 & 20.81 $\pm$ 0.88 & 23.62 $\pm$ 0.89 & 22.23 $\pm$ 0.84 & 25.04 $\pm$ 0.85\\
MSTGCN\cite{guo19} & 21.15 $\pm$ 0.72 & 23.99 $\pm$ 0.70 & 24.51 $\pm$ 0.57 & 27.32 $\pm$ 0.56 & 26.10 $\pm$ 0.24 & 28.90 $\pm$ 0.26\\
ASTGCN\cite{guo19} & 20.20 $\pm$ 0.81 & 23.15 $\pm$ 0.83 & 24.80 $\pm$ 0.33 & 27.66 $\pm$ 0.38 & 26.87 $\pm$ 0.33 & 29.69 $\pm$ 0.38\\
$\rm{PM_{2.5}GNN}$\cite{wang20} & 16.41 $\pm$ 0.74 & 19.44 $\pm$ 0.75 & 19.46 $\pm$ 0.63 & 22.63 $\pm$ 0.63 & 21.21 $\pm$ 0.59 & 24.49 $\pm$ 0.60\\
GAGNN\cite{chen21} & 19.70 $\pm$ 0.43 & 22.69 $\pm$ 0.45 & 21.15 $\pm$ 0.49 & 26.25 $\pm$ 0.50 & 25.46 $\pm$ 0.47 & 28.87 $\pm$ 0.45\\
HighAir\cite{xu21} & 16.53 $\pm$ 0.89 & 19.80 $\pm$ 0.87 & 20.22 $\pm$ 0.73 & 23.22 $\pm$ 0.75 & 21.36 $\pm$ 0.64 & 24.65 $\pm$ 0.65\\
\midrule
\textbf{M2G2} & \textbf{15.39 $\pm$ 0.46} & \textbf{18.05 $\pm$ 0.44} & \textbf{18.17 $\pm$ 0.58} & \textbf{21.12 $\pm$ 0.56} & \textbf{19.15 $\pm$ 0.65} & \textbf{21.93 $\pm$ 0.63}\\

\bottomrule
\label{baselinepm25}
\end{tabular}}
\end{table}

In this section, we compare the MAE and RMSE metrics of our model and the comparison baselines. To ensure the repeatability of the experiments and the stability of the results, we run each method five times with various seeds to determine the mean value and standard deviation. In order to more effectively illustrate the experimental results, we provide the forecasts for the upcoming times in segments: 1-24 hours, 25-48 hours, and 49-72 hours. These results are shown in Table. \ref{baselinepm25}, and it can be seen that all segments of our model outperform the comparative methods. In terms of MAE, RMSE, we do better than the second best method (i.e. $\rm{PM_{2.5}GNN}$) by (6.22\%, 6.63\%, 9.71\%) and (7.72\%, 6.67\%, 10.45\%), respectively. It can be observed that M2G2 demonstrates relatively minimal deterioration in long-term forecasts (49-72 hours) and continues to maintain a favorable scale of predictive accuracy compared to other approaches.

\begin{figure}[htbp]
\centering
\subfigure[3 hour prediction horizon]{
\includegraphics[width=7.5cm]{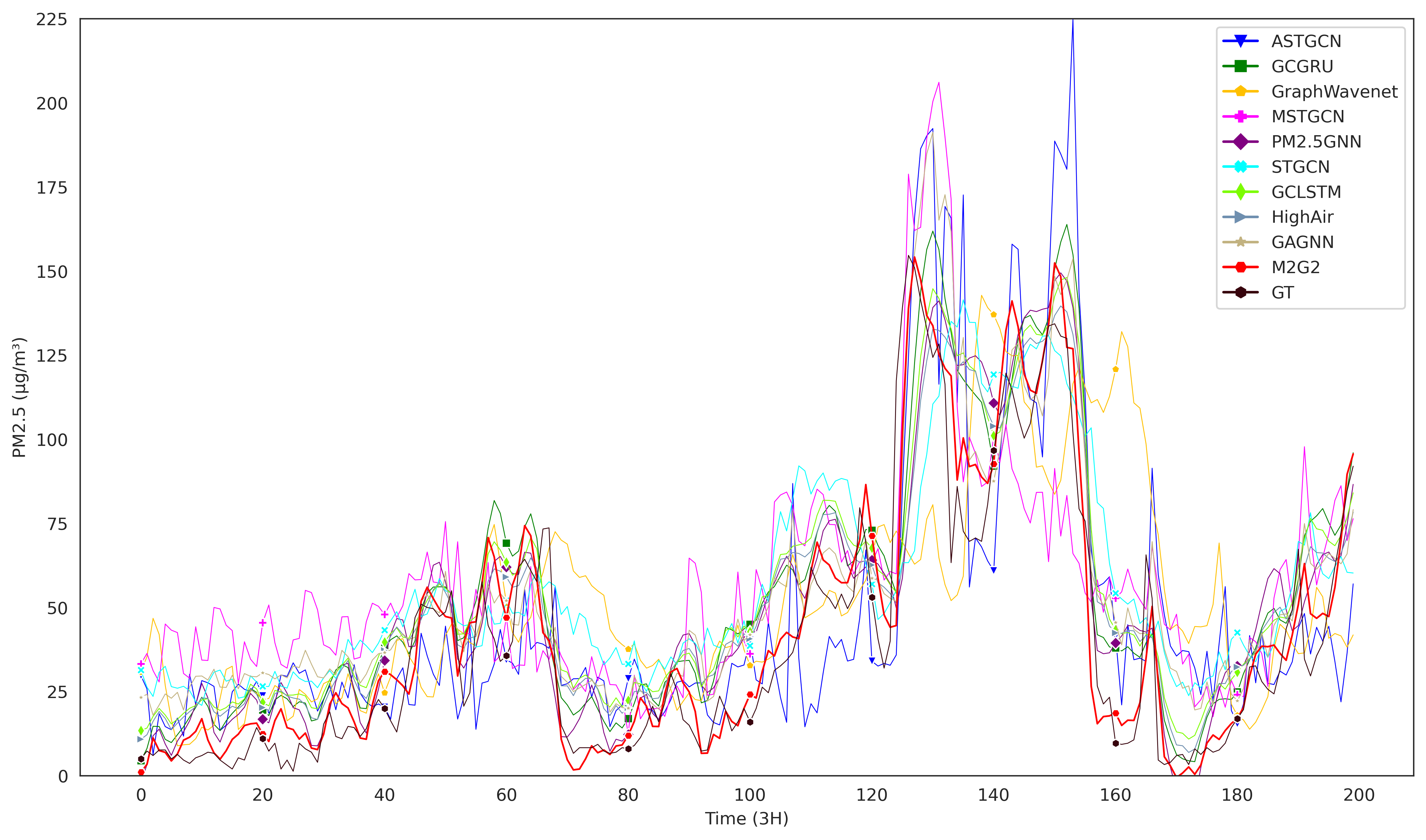}
}
\quad
\subfigure[24 hour prediction horizon]{
\includegraphics[width=7.5cm]{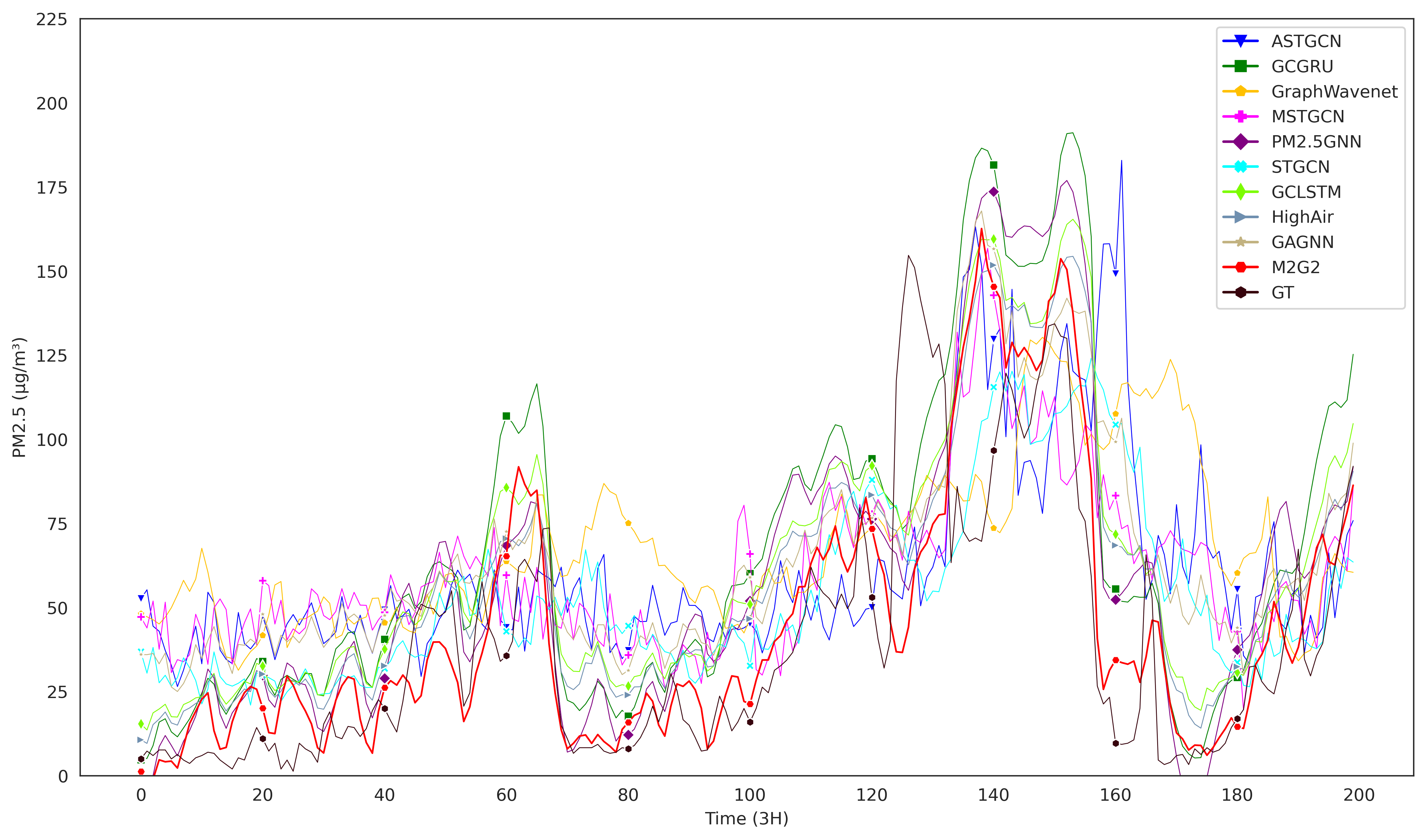}
}
\quad
\subfigure[48 hour prediction horizon]{
\includegraphics[width=7.5cm]{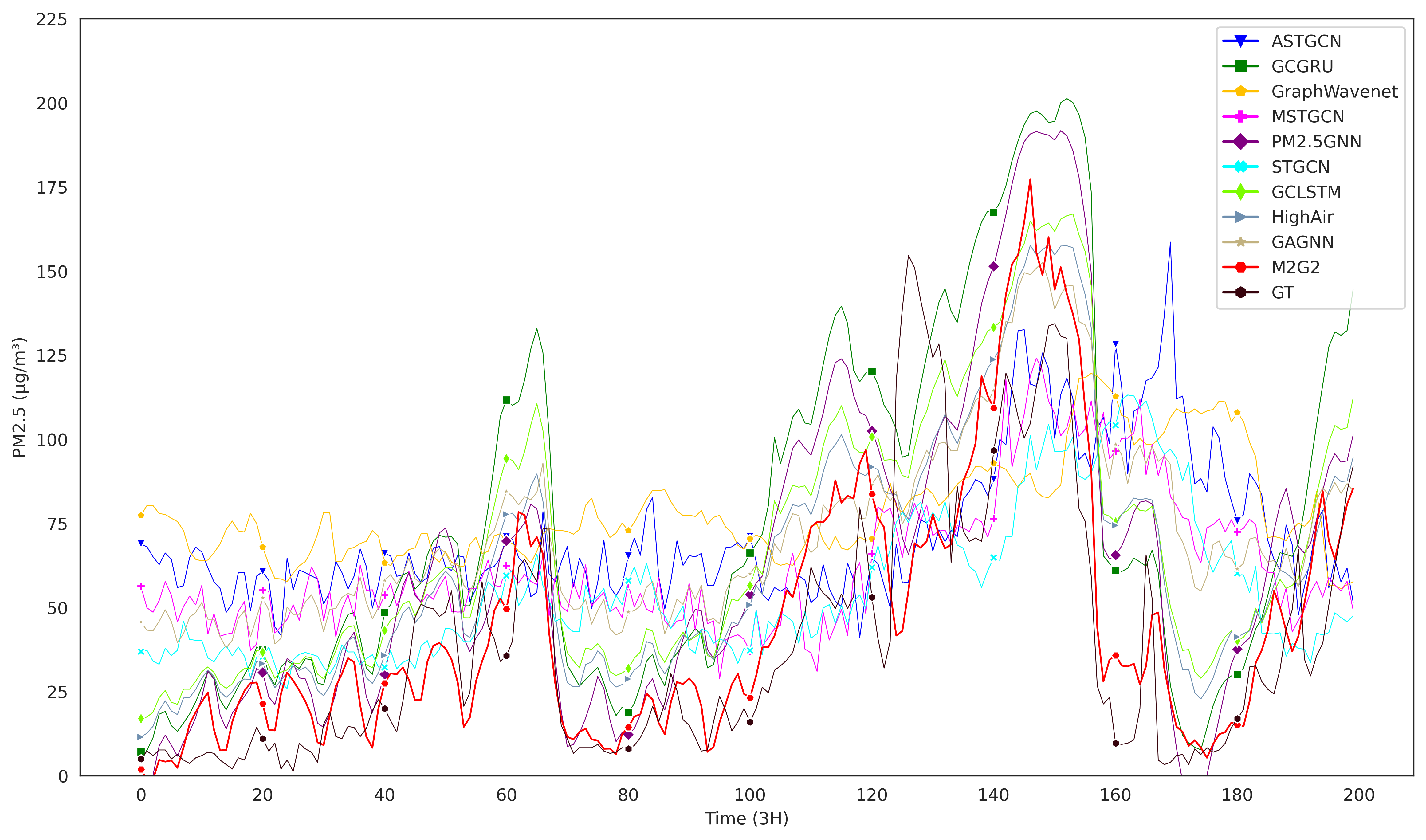}
}
\quad
\subfigure[72 hour prediction horizon]{
\includegraphics[width=7.5cm]{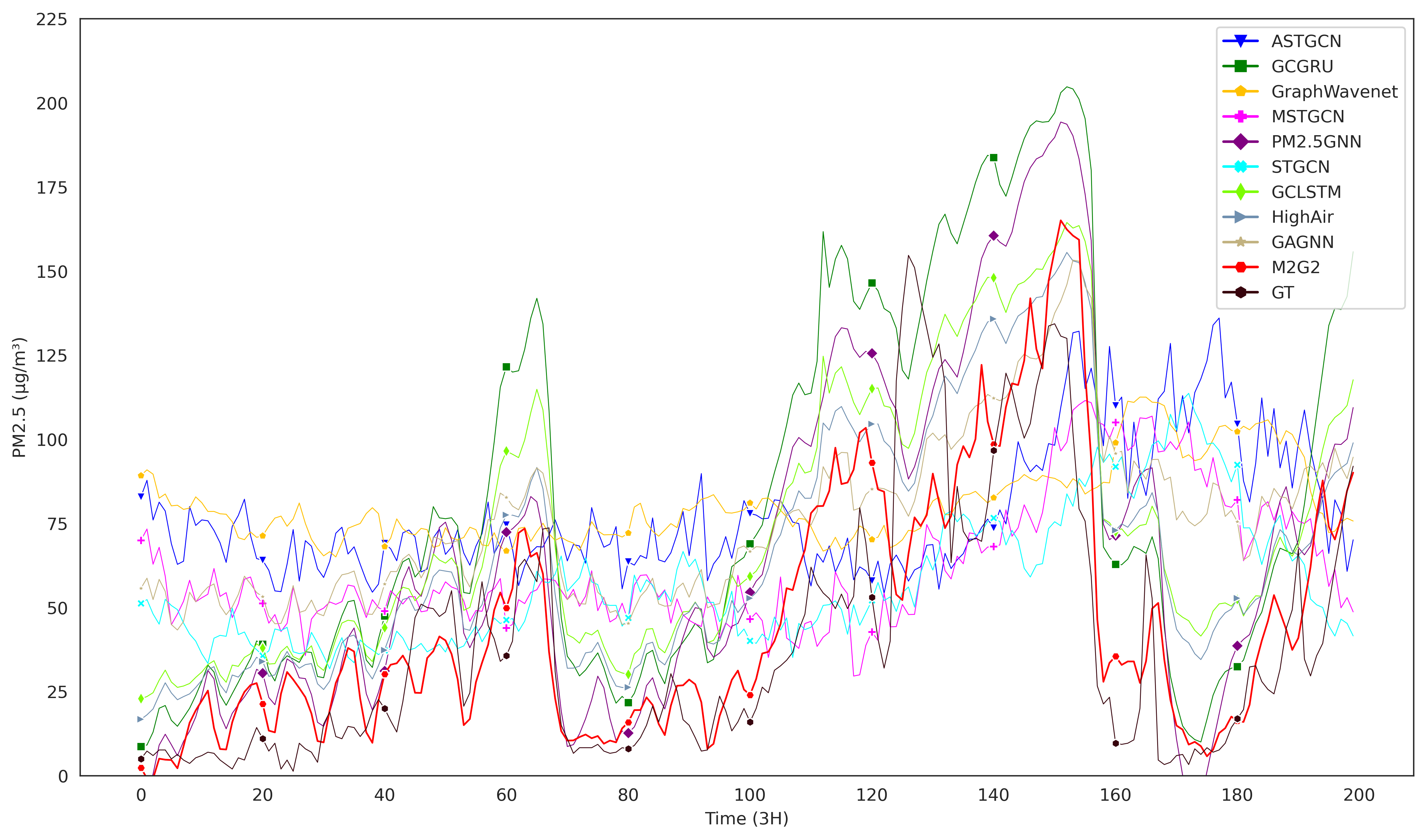}
}
\caption{The prediction for different hour prediction horizons. Subfigure (a) represents the results of predicting pollutant concentrations for the next 3 hours, while (b) to (d) correspond to the next 24 hours, 48 hours, and 72 hours, respectively.}
\label{linechart0}
\end{figure}

\begin{figure}[htbp]
\centering
\includegraphics[width=0.9\textwidth]{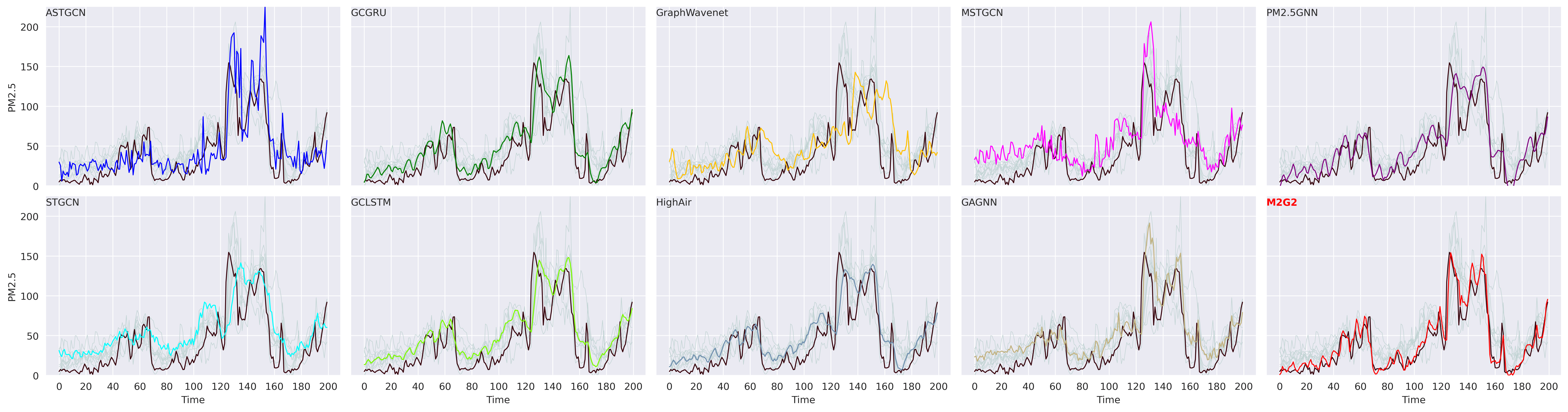}
\caption{The fine-grained comparison for 3 hour prediction horizon. The black line represents the ground truth, while the gray lines indicate all techniques save the current one. The remaining colored line reflects the method that corresponds to the current subplot.}
\label{linechart1}
\end{figure}

\begin{figure}[htbp]
\centering
\includegraphics[width=0.9\textwidth]{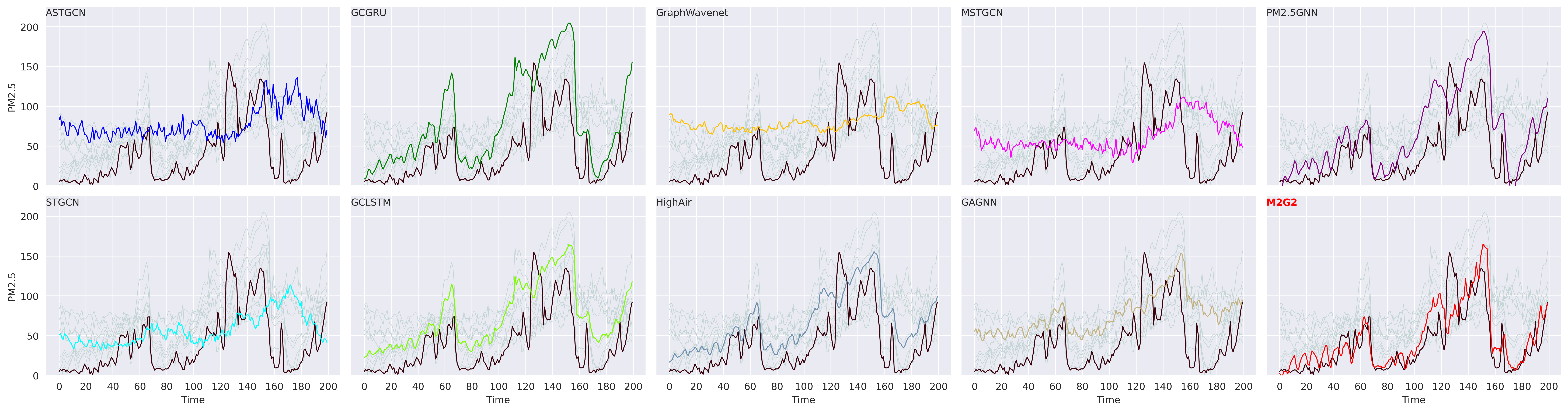}
\caption{The fine-grained comparison for 72 hour prediction horizon. The black line represents the ground truth, while the gray lines indicate all techniques save the current one. The remaining colored line reflects the method that corresponds to the current subplot.}
\label{linechart2}
\end{figure}

\begin{figure}[H]
\centering
\includegraphics[width=0.95\textwidth]{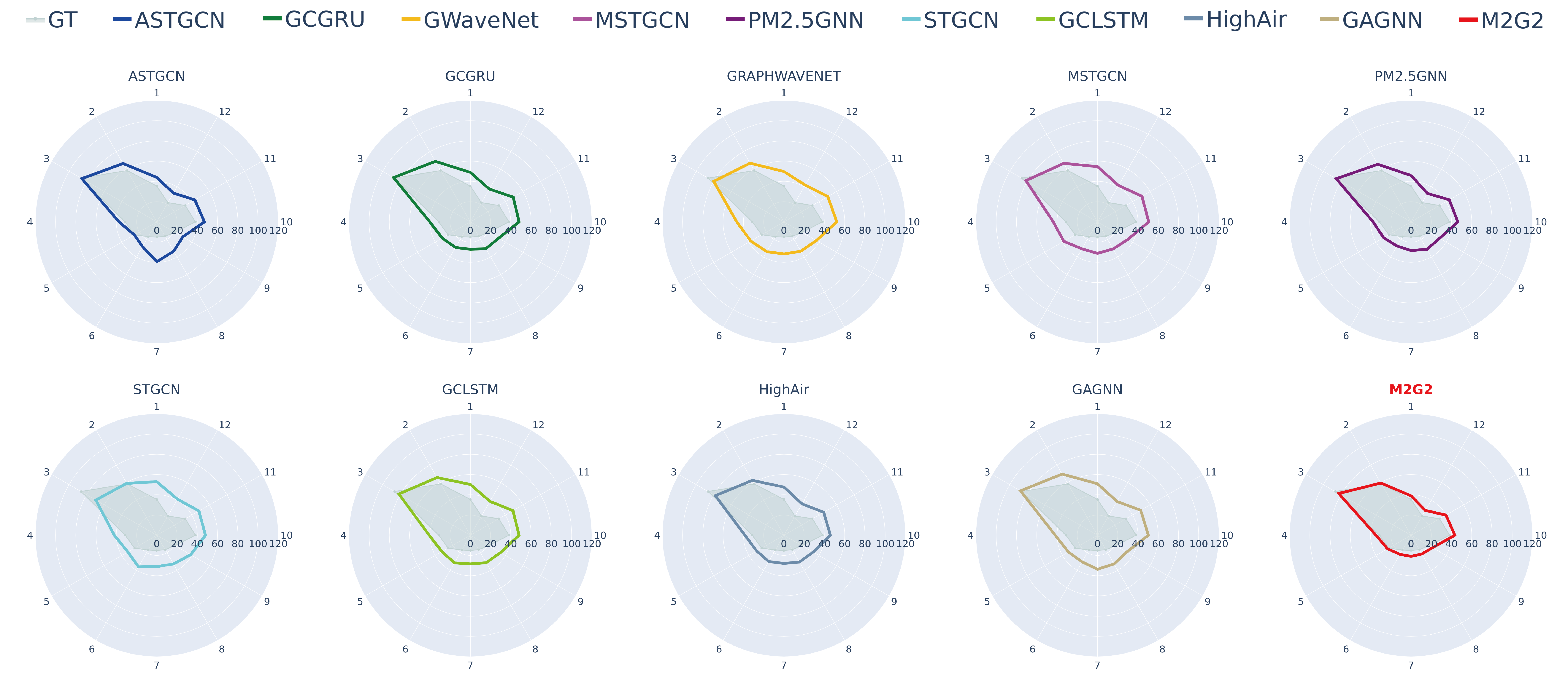}
\caption{The average value of each month for 3 hour prediction horizon. The gray background region represents the ground truth of air pollutant concentration. The circular regions are labeled in counterclockwise order as 1, 2, 3...12, representing the twelve months. The distance from the center of the circle to the position of the folded line in the direction of each month is the monthly average of the pollutant concentration predicted by the current method. Comparing this value with the ground truth represented by the gray background reflects the performance of the respective method.}
\label{radar0}
\end{figure}

\vspace{+5.em}

\begin{figure}[H]
\centering
\includegraphics[width=0.99\textwidth]{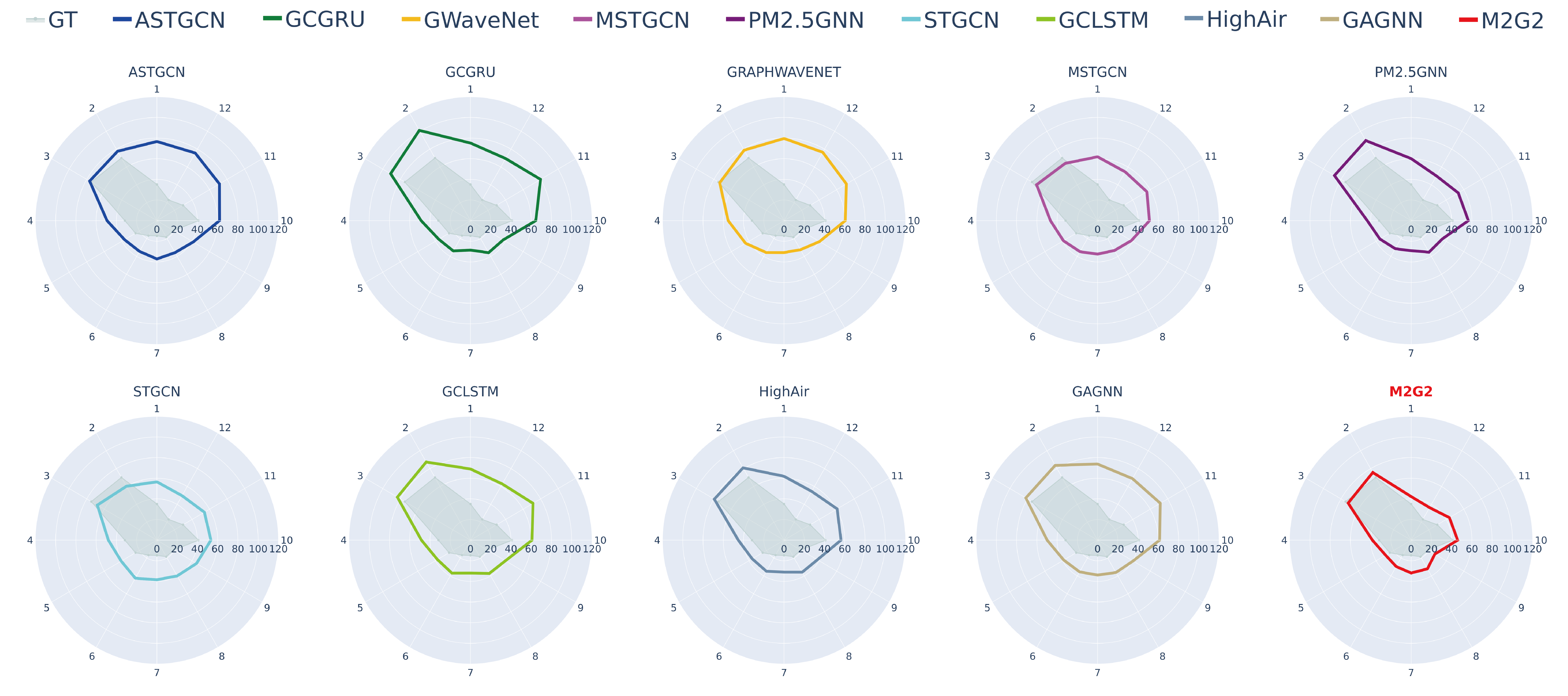}
\caption{The average value of each month for 72 hour prediction horizon. The gray background region represents the ground truth of air pollutant concentration. The circular regions are labeled in counterclockwise order as 1, 2, 3...12, representing the twelve months. The distance from the center of the circle to the position of the folded line in the direction of each month is the monthly average of the pollutant concentration predicted by the current method. Comparing this value with the ground truth represented by the gray background reflects the performance of the respective method.}
\label{radar1}
\end{figure}


Fig. \ref{linechart0} displays a randomly chosen subset of observation stations and time periods for visualization purposes. The time axis in the figure is measured in 3-hour intervals. To more accurately compare the methods, we provide prediction curves for different prediction horizons. Our proposed model demonstrates superior performance across all prediction horizons, even when predicting ${\rm PM}_{2.5}$ concentrations up to 72 hours in advance. This suggests that our model is capable of capturing the complex variations in air pollutant concentrations over time, which is critical for accurate air quality forecasting. 

\vspace{+0.5em}

Furthermore, Fig. \ref{linechart1} and \ref{linechart2} provide a more detailed view of the performance differences between our proposed method and the comparative models. In particular, these figures highlight the superior fine-grained prediction accuracy of our method. As shown in Fig. \ref{linechart1}, which depicts a 3-hour prediction horizon, the second-best model $\rm{PM_{2.5}GNN}$ is unable to capture the significant fluctuations in ${\rm PM}_{2.5}$ concentration that occur in the time axis range of 125–175, whereas our method achieves a much better fit. When extending the prediction horizon to 72 hours, as shown in Fig. \ref{linechart2}, our proposed method again outperforms the other methods, achieving much closer agreement with the actual ${\rm PM}_{2.5}$ concentration across almost all time periods. These results demonstrate the superior performance of our method in accurately predicting air pollutant concentrations over a range of time horizons.

\vspace{+0.5em}

While the above line graph showcases the fine-grained prediction accuracy of our method, the seasonal variation in ${\rm PM}_{2.5}$ concentration is also an important factor to consider. To illustrate our model's performance on a larger time scale, we present Figs. \ref{radar0} and \ref{radar1}, which depict the monthly mean values of the predicted and actual pollutant concentrations. The lines in the figures represent the mean values of the predicted pollutant concentrations for the corresponding models, while the green filled box represents the mean values of the actual observations. In particular, our proposed method shows excellent performance during the winter months, when the ${\rm PM}_{2.5}$ concentration is typically high. As shown in Fig. \ref{radar0}, our model outperforms the second-best model $\rm{PM_{2.5}GNN}$ in the months with low pollutant concentrations, particularly for a 3-hour prediction horizon. Moreover, in Fig. \ref{radar1}, which depicts a 72-hour prediction horizon, the other comparative methods exhibit relatively large deviations from the actual monthly mean ${\rm PM}_{2.5}$ concentrations, while our model still maintains a similar shape to the actual observations. These results demonstrate the superior performance of our proposed method for the long-term prediction of air pollutant concentrations, particularly in the presence of seasonal variations.

\vspace{+0.5em}

\vspace{+0.5em}

\begin{figure}[H]
\centering
\includegraphics[width=0.99\textwidth]{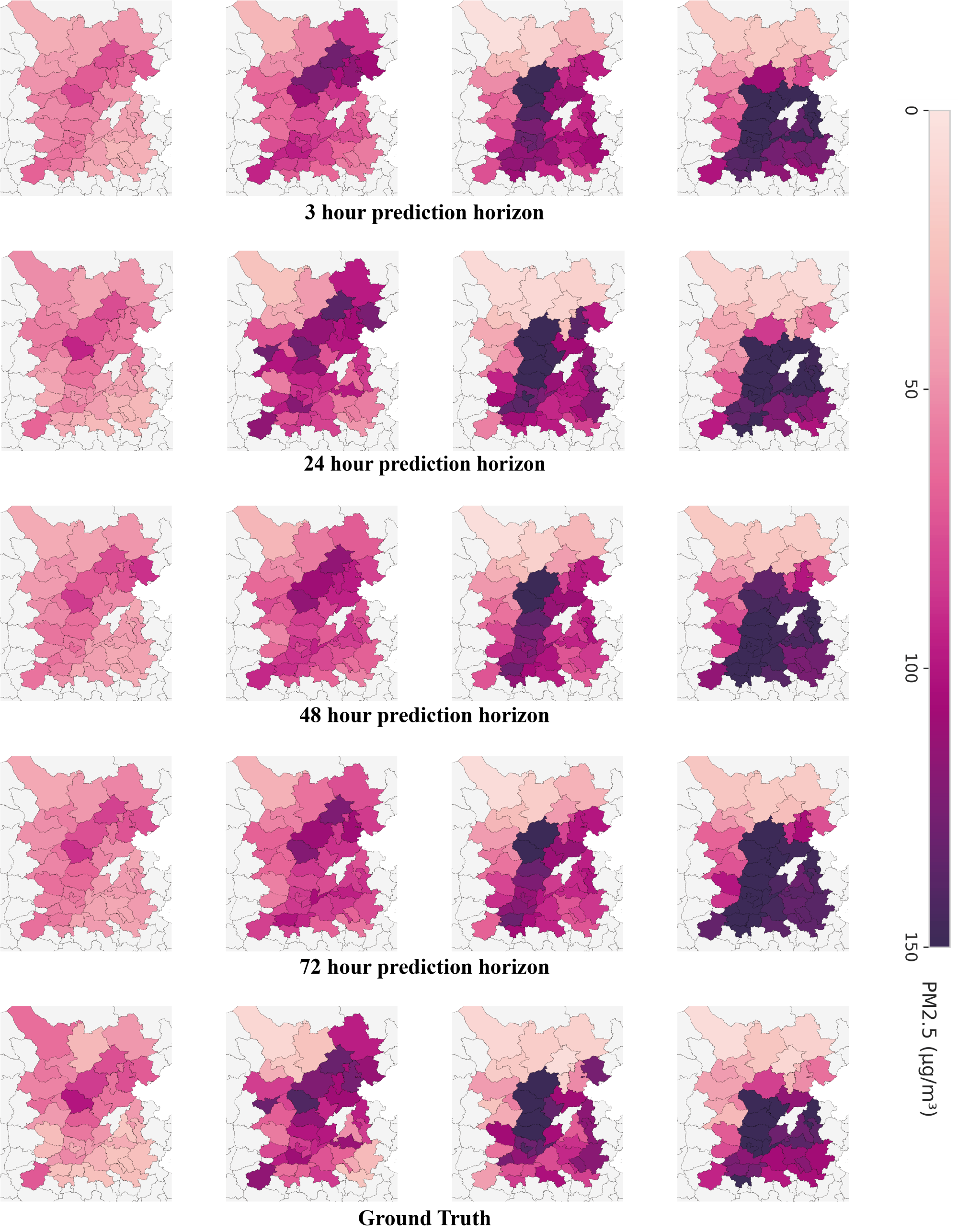}
\caption{The choropleth map of the prediction concentrations and the ground truth. The shades of color represent the values of pollutant concentrations, with darker shades indicating more severe pollution. The specific correspondence can be referred to the colorbar on the right side. The first row represents the predicted pollutant concentration for the next 3 hours, and so on. The last row represents the actual values of pollutant concentrations.}

\label{choroplethMap}
\end{figure}

To provide a regional perspective on the performance of our proposed method, we present Fig. \ref{choroplethMap}, which illustrates the predicted and actual ${\rm PM}_{2.5}$ concentrations for various regions. The upper row of the figure displays the prediction results of our M2G2 model for different prediction horizons, while the lower row shows the actual ${\rm PM}_{2.5}$ concentrations. Notably, our model achieves high accuracy at the fine-grained spatial scale, with correct predictions made for both high- and low-concentration locations, as shown in the 3-hour prediction horizon. Furthermore, as the prediction horizon increases, our M2G2 model maintains superior performance, demonstrating its stability in forecasting air pollutant concentrations over longer time periods. These results highlight the effectiveness of our proposed method in capturing the spatial patterns of air pollutant concentrations across different regions.

\subsection{Effectiveness of MT-GRU}

\begin{figure}[H]
\centering
\includegraphics[width=0.99\textwidth]{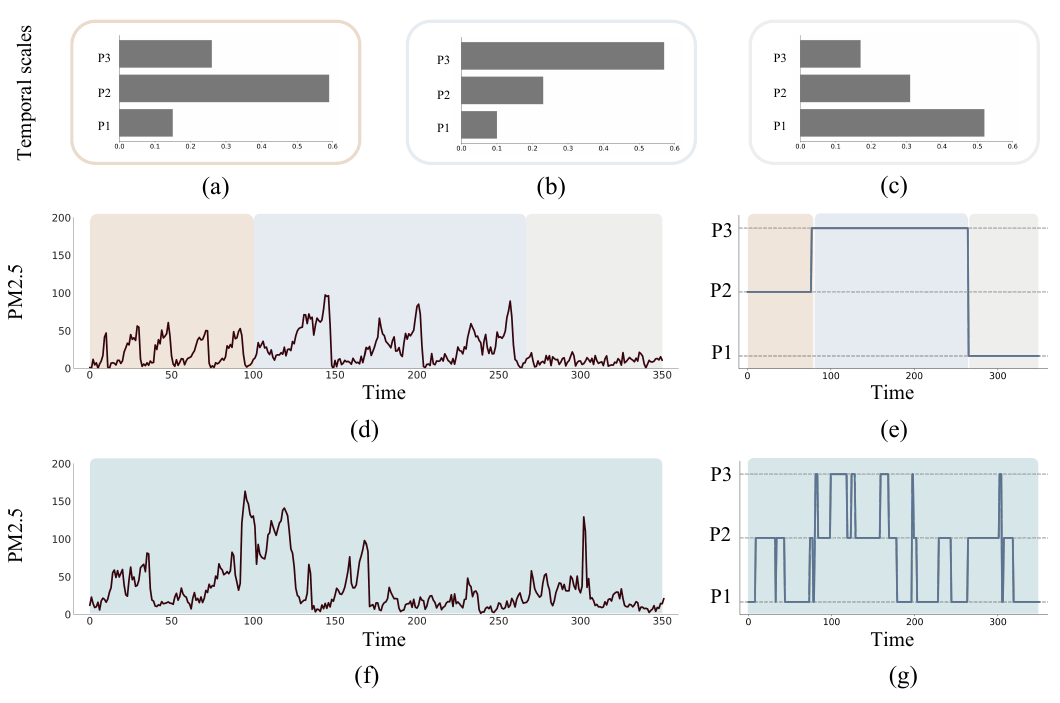}
\caption{P1, P2, and P3 represent temporal scales in ascending order, indicating different periods for updating the GRU hidden state. They take values of 1, 2, and 4, respectively. Subfigures (a), (b), and (c) correspond to different time intervals in (d): red background, blue background, and gray background. The red interval approximately spans the range 0-100 on the x-axis, the blue interval spans 100-260, and the gray interval spans 260-350. They correspond to sequences with medium period, long period, and short period, respectively. Similarly, subfigures (a), (b), and (c) use red, blue, and gray boxes to indicate different periodic sequences. The x-axis is represented by the dynamic temporal scale weights calculated by formula \ref{dynamicW}, which indicates the importance of each temporal scale for prediction according to the MT-GRU model. In subfigure (d), the time series is artificially generated and exhibits noticeable period differences, highlighting the ability of MT-GRU to timely perceive prominent scale features at each time step. Subfigure (f) presents a real-world sequence, further validating the effectiveness of MT-GRU in practical scenarios. Subfigures (e) and (g) depict the temporal scale that carries the highest weight at each time step.}

\label{effectiveMT}
\end{figure}

To rigorously validate the MT-GRU module's ability to capture features from different time periods, we visualize the characteristics of dynamic temporal scale weights in Fig. \ref{effectiveMT}, where P1, P2, and P3 correspond to ascending temporal scales. Subfigure (d) presents an artificially generated time series, with red, blue, and gray backgrounds representing sequences of different periods: medium, long, and short, respectively. The y-axis in subfigure (e) indicates the temporal scale with the highest dynamic weight at the current time step. In the testing of artificial data, we observe that MT-GRU effectively learns the most prominent periodic scale. Furthermore, in subfigures (a) to (c), we conduct additional sampling of specific time steps from different periodic sub-sequences in subfigure (d) to visualize the corresponding dynamic weight values. Subfigures (a) to (c) correspond to the sampling of the medium-period sequence (red), long-period sequence (blue), and short-period sequence (gray), respectively. It can be observed that the weight distribution exhibits high distinctiveness and correctly corresponds to the respective period, thereby validating the reliability of MT-GRU. 

Moreover, in the case of real data (subfigure f), MT-GRU demonstrates its capability to track the current prominent period. As shown on the time axis, the range 60-160 clearly represents a larger periodic scale, while 200-250 represents a smaller periodic scale. The corresponding time intervals in subfigure (g) demonstrate that MT-GRU can learn the features of the dominant time scale at the given moments.

\subsection{Ablation Study}

Ablation Study is conducted to examine the significance or contribution of specific components or factors within a system. By removing certain components and modules and observing the resulting changes in model performance, we can validate the roles of these components or modules. Additionally, for certain adjustable model hyperparameters, we perform experiments with various settings to test the model's performance limits and sensitivity. Subsequently, we will gradually validate the effects of MS-GCN, effects of MT-GRU, and effects of choice of temporal scale.

\vspace{+1.em}
\begin{figure}[H]
\centering
\includegraphics[width=0.6\textwidth]{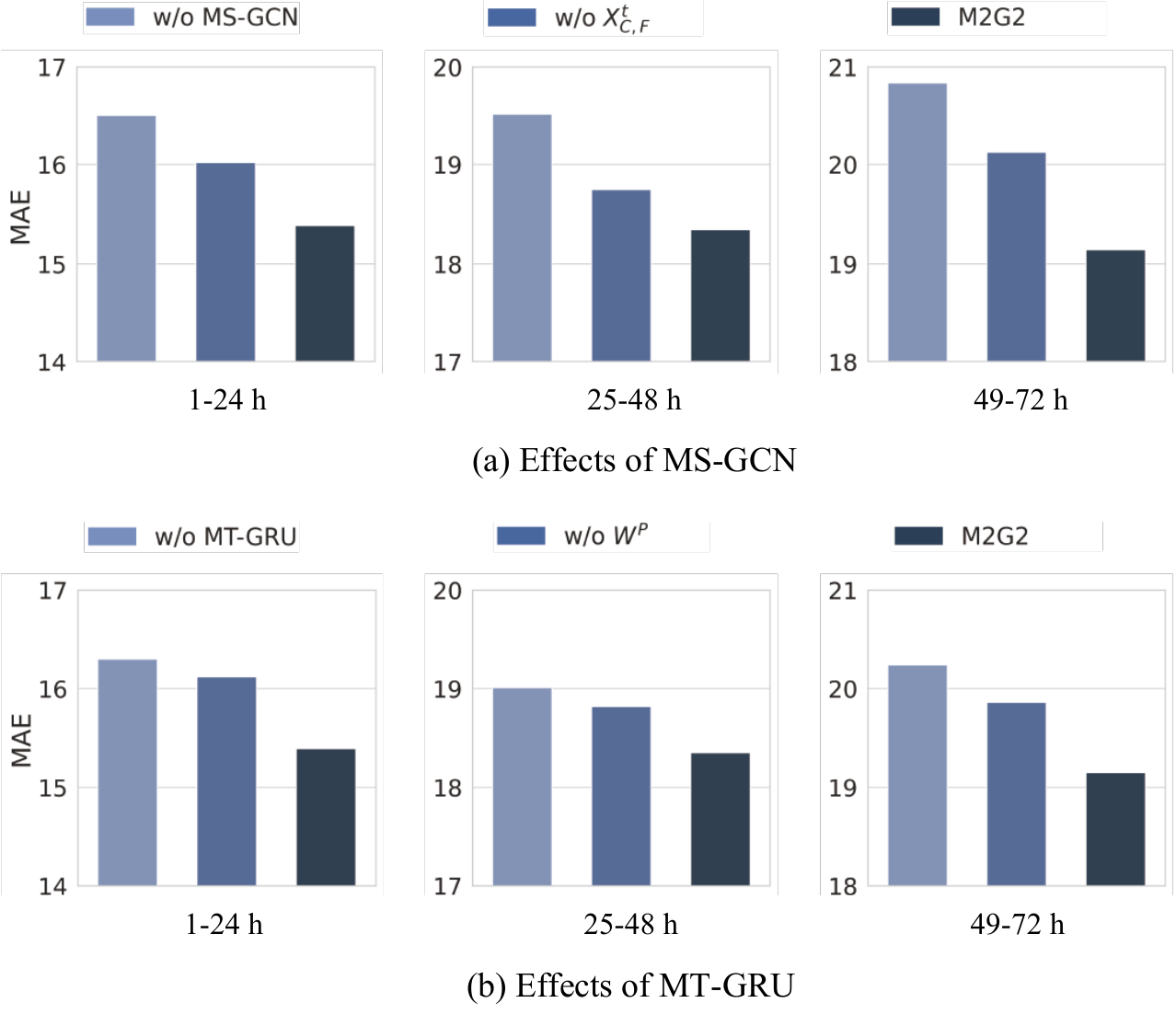}
\caption{Ablation Study on MS-GCN and MT-GRU respectively. The y-axis represents the MAE (Mean Absolute Error) values. 1-24h, 25-48h, and 49-72h respectively indicate the performance for three different prediction time ranges: 1-24 hours, 25-48 hours, and 49-72 hours.}

\label{AS}
\end{figure}

\subsubsection{Effects of MS-GCN}

To verify the validity of multiple spatial scales, we design several spatial scale related variants to compare with our model. a) \textbf{w/o MS-GCN}: we remove all components of the city-scale. That is, we rely only on the station-scale for prediction. b) \textbf{w/o $\bm{X}_{C,F}^t$}: Similarly, we remove the mechanism for station features to be transferred to the city-scale (Eq. \ref{s2cFusion}). According to the results shown in Fig. \ref{AS}(a), the introduction of city-scale will reduce the MAE and RMSE, and increase the accuracy of the prediction. Moreover, the transfer of station-scale features to city-scale is also indirectly beneficial.

\subsubsection{Effects of MT-GRU}

In addition to studying the role of MS-GCN, it is important to assess the impact of MT-GRU on the overall performance of our proposed method. To this end, we conduct two ablation experiments as follows: (a) \textbf{w/o MT-GRU}: We remove the MT-GRU component and replace it with the original GRU for temporal modeling; (b) \textbf{w/o $ \bm{W}^P $}: We eliminate the dynamic weight generation process in MT-GRU, whereby dynamic weights $ \bm{W}^P $ are established for each temporal scale (as shown in Eq. \ref{dynamicW}), and instead use consistent weights for all temporal scales. As illustrated in Fig. \ref{AS}(b), the results demonstrate that the design of MT-GRU can significantly improve the prediction performance, and the dynamic weights play a crucial role in achieving this improvement.

\subsubsection{Effects of choice of temporal scale}

\begin{table}[htbp]
\caption{The effect of the choice of different temporal scales on prediction accuracy. Here the value in temporal scale vector ${\bm P}$ denotes different update steps, with each step being 3 hours. For example, $[1, 2]$ indicates that the actual update steps are $[3 hours, 6 hours]$.}
\vspace{1.em}
\centering
\setlength{\tabcolsep}{5.4mm}{
\begin{tabular}{cccccc}
\toprule
{} & \textbf{\makecell[c]{Temporal Scale \\Vector ${\bm P}$}}& \textbf{Metric} & \textbf{1-24h} & \textbf{25-48h} & \textbf{49-72h}\\
\midrule
\multirow{6}*{\textbf{2 hidden states}} & \multirow{2}*{[1, 2]} & MAE & 16.30 & 19.43 & 20.48\\
{} & {} & RMSE & 19.05 & 22.19 & 23.24\\
\cmidrule{3-6}
{} & \multirow{2}*{[1, 4]} & MAE & 16.10 & 19.06 & 19.97\\
{} & {} & RMSE & 18.89 & 21.85 & 22.76\\
\cmidrule{3-6}
{} & \multirow{2}*{[1, 8]} & MAE & 16.31 & 19.33 & 20.33\\
{} & {} & RMSE & 19.10 & 22.12 & 23.12\\
\midrule
\multirow{6}*{\textbf{3 hidden states}} & \multirow{2}*{[1, 2, 4]} & MAE & \textbf{15.39} & \textbf{18.35} & \textbf{19.15}\\
{} & {} & RMSE & \textbf{18.15} & \textbf{21.12} & \textbf{21.93}\\
\cmidrule{3-6}
{} & \multirow{2}*{[1, 2, 8]} & MAE & 15.71 & 18.68 & 19.60\\
{} & {} & RMSE & 18.47 & 21.46 & 22.39\\
\cmidrule{3-6}
{} & \multirow{2}*{[1, 4, 8]} & MAE & 15.72 & 18.58 & 19.39\\
{} & {} & RMSE & 18.50 & 21.36 & 22.18\\
\midrule
\multirow{2}*{\textbf{4 hidden states}} & \multirow{2}*{[1, 2, 4, 8]} & MAE & 16.03 & 19.24 & 20.41\\
{} & {} & RMSE & 18.79 & 22.00 & 23.17\\
\bottomrule
\label{choiceP}
\end{tabular}}
\end{table}

To further explore the effectiveness of the MT-GRU, we investigate the impact of the temporal scale vector $ \bm{P} $, which represents different GRU hidden state update periods. According to Table. \ref{choiceP} , the number of temporal scales $ |\bm{P}| $ selected as 3 is superior than others. The optimum upper temporal scale is supposed to be 4, with an increase to 8 having a detrimental effect due to excessive redundancy.

\subsection{Experimentation of other pollutant indexes}

As described in \ref{section: dataset}, the dataset contains four different air pollutants: ${\rm PM}_{2.5}$, ${\rm PM}_{10}$, ${\rm NO}_2$, ${\rm O}_3$, and we also conducted comparison tests for pollutants other than ${\rm PM}_{2.5}$, the results of which are shown in Table. \ref{baselinepm10}, \ref{baselineno2} and \ref{baselineo3}. We outperformed all other air quality indicators, demonstrating the generalizability and applicability of our model to the problem of spatiotemporal prediction of air quality. Furthermore, it is once more confirmed that designing spatial and temporal multi-scale components is essential in the objective world.

\vspace{-1.em}

\begin{table}[H]
\caption{${\rm PM}_{10}$ baseline table. 1-24h, 25-48h, and 49-72h represent the performance of predicting pollutant concentrations for the next 1-24 hours, 25-48 hours, and 49-72 hours, respectively.}
\centering
\resizebox{\textwidth}{!}{
\begin{tabular}{ccccccc}
\toprule
\multirow{2}*{Model} & \multicolumn{2}{c}{\textbf{\underline{\qquad\quad 1-24h\quad\qquad}}} & \multicolumn{2}{c}{\textbf{\underline{\quad\qquad 25-48h\qquad\quad}}} & \multicolumn{2}{c}{\textbf{\underline{\quad\qquad 49-72h\qquad\quad}}} \\
&MAE&RMSE&MAE&RMSE&MAE&RMSE\\
\midrule
GCGRU\cite{li18} & 36.17 $\pm$ 1.02 & 43.44 $\pm$ 1.07 & 42.51 $\pm$ 1.18 & 49.83 $\pm$ 1.20 & 44.87 $\pm$ 1.39 & 52.18 $\pm$ 1.39\\
STGCN\cite{yu18} & 41.38 $\pm$ 0.47 & 48.80 $\pm$ 0.47 & 47.12 $\pm$ 0.71 & 54.70 $\pm$ 0.64 & 49.54 $\pm$ 0.81 & 57.09 $\pm$ 0.26\\
GWNET\cite{wu19} & 47.02 $\pm$ 0.35 & 54.46 $\pm$ 0.38 & 49.68 $\pm$ 0.63 & 57.04 $\pm$ 0.66 & 50.88 $\pm$ 0.60 & 58.18 $\pm$ 0.57\\
GCLSTM\cite{qi19} & 37.05 $\pm$ 1.31 & 44.33 $\pm$ 1.32 & 43.53 $\pm$ 1.08 & 50.86 $\pm$ 1.05 & 45.92 $\pm$ 1.35 & 53.25 $\pm$ 1.32\\
MSTGCN\cite{guo19} & 41.21 $\pm$ 1.14 & 48.76 $\pm$ 1.09 & 46.82 $\pm$ 0.83 & 54.35 $\pm$ 0.84 & 49.24 $\pm$ 1.24 & 56.74 $\pm$ 1.25\\
ASTGCN\cite{guo19} & 38.67 $\pm$ 0.83 & 46.11 $\pm$ 0.87 & 47.25 $\pm$ 0.59 & 54.68 $\pm$ 0.60 & 50.59 $\pm$ 0.24 & 57.99 $\pm$ 0.26\\
$\rm{PM_{2.5}GNN}$\cite{wang20} & 35.79 $\pm$ 0.83 & 42.91 $\pm$ 0.85 & 41.82 $\pm$ 0.88 & 49.11 $\pm$ 0.86 & 45.04 $\pm$ 1.07 & 52.27 $\pm$ 1.05\\
GAGNN\cite{chen21} & 39.20 $\pm$ 0.91 & 46.73 $\pm$ 0.92 & 45.42 $\pm$ 0.55 & 52.55 $\pm$ 0.52 & 47.57 $\pm$ 0.87 & 54.65 $\pm$ 0.84\\
HighAir\cite{xu21} & 35.82 $\pm$ 1.22 & 43.06 $\pm$ 1.24 & 41.94 $\pm$ 0.51 & 49.25 $\pm$ 0.53 & 44.69 $\pm$ 0.73 & 51.97 $\pm$ 0.67\\
\midrule
\textbf{M2G2} & \textbf{33.72 $\pm$ 0.87} & \textbf{40.15 $\pm$ 0.84} & \textbf{39.51 $\pm$ 1.02} & \textbf{46.32 $\pm$ 0.99} & \textbf{41.32 $\pm$ 1.09} & \textbf{48.23 $\pm$ 1.07}\\
\bottomrule
\label{baselinepm10}
\end{tabular}}
\vspace{-1.em}
\end{table}

\vspace{-1.em}

\begin{table}[H]
\caption{${\rm NO}_2$ baseline table. 1-24h, 25-48h, and 49-72h represent the performance of predicting pollutant concentrations for the next 1-24 hours, 25-48 hours, and 49-72 hours, respectively.}
\centering
\resizebox{\textwidth}{!}{ 
\begin{tabular}{ccccccc}
\toprule
\multirow{2}*{Model} & \multicolumn{2}{c}{\textbf{\underline{\qquad\quad 1-24h\quad\qquad}}} & \multicolumn{2}{c}{\textbf{\underline{\quad\qquad 25-48h\qquad\quad}}} & \multicolumn{2}{c}{\textbf{\underline{\quad\qquad 49-72h\qquad\quad}}} \\
&MAE&RMSE&MAE&RMSE&MAE&RMSE\\
\midrule
GCGRU\cite{li18} & 9.31 $\pm$ 0.59 & 11.16 $\pm$ 0.61 & 10.31 $\pm$ 0.55 & 12.19 $\pm$ 0.56 & 10.84 $\pm$ 0.63 & 12.72 $\pm$ 0.63\\
STGCN\cite{yu18} & 11.79 $\pm$ 0.15 & 13.85 $\pm$ 0.15 & 13.23 $\pm$ 0.23 & 15.34 $\pm$ 0.24 & 13.66 $\pm$ 0.23 & 15.77 $\pm$ 0.22\\
GWNET\cite{wu19} & 14.28 $\pm$ 0.06 & 16.39 $\pm$ 0.06 & 14.59 $\pm$ 0.09 & 16.71 $\pm$ 0.09 & 14.71 $\pm$ 0.06 & 16.82 $\pm$ 0.06\\
GCLSTM\cite{qi19} & 9.22 $\pm$ 0.20 & 10.99 $\pm$ 0.21 & 10.38 $\pm$ 0.25 & 12.18 $\pm$ 0.25 & 11.11 $\pm$ 0.22 & 12.92 $\pm$ 0.22\\
MSTGCN\cite{guo19} & 11.68 $\pm$ 0.27 & 13.76 $\pm$ 0.27 & 13.18 $\pm$ 0.15 & 15.30 $\pm$ 0.15 & 13.48 $\pm$ 0.10 & 15.59 $\pm$ 0.10\\
ASTGCN\cite{guo19} & 11.22 $\pm$ 0.15 & 13.27 $\pm$ 0.14 & 12.90 $\pm$ 0.03 & 14.99 $\pm$ 0.03 & 13.37 $\pm$ 0.15 & 15.47 $\pm$ 0.14\\
$\rm{PM_{2.5}GNN}$\cite{wang20} & 9.07 $\pm$ 0.51 & 10.84 $\pm$ 0.52 & 10.17 $\pm$ 0.75 & 11.97 $\pm$ 0.74 & 10.85 $\pm$ 1.26 & 12.65 $\pm$ 1.24\\
GAGNN\cite{chen21} & 10.59 $\pm$ 0.59 & 13.67 $\pm$ 0.58 & 11.56 $\pm$ 0.66 & 14.26 $\pm$ 0.64 & 12.28 $\pm$ 0.45 & 14.91 $\pm$ 0.44\\
HighAir\cite{xu21} & 9.33 $\pm$ 0.24 & 11.16 $\pm$ 0.24 & 10.25 $\pm$ 0.35 & 12.12 $\pm$ 0.36 & 10.70 $\pm$ 0.38 & 12.58 $\pm$ 0.39\\
\midrule
\textbf{M2G2} & \textbf{8.58 $\pm$ 0.68} & \textbf{10.29 $\pm$ 0.67} & \textbf{9.18 $\pm$ 0.64} & \textbf{11.04 $\pm$ 0.71} & \textbf{8.74 $\pm$ 0.76} & \textbf{10.55 $\pm$ 0.74}\\
\bottomrule
\label{baselineno2}
\end{tabular}}
\vspace{-1.em}
\end{table}

\vspace{-1.em}

\begin{table}[H]
\caption{${\rm O}_3$ baseline table. 1-24h, 25-48h, and 49-72h represent the performance of predicting pollutant concentrations for the next 1-24 hours, 25-48 hours, and 49-72 hours, respectively.}
\centering
\resizebox{\textwidth}{!}{ 
\begin{tabular}{ccccccc}
\toprule
\multirow{2}*{Model} & \multicolumn{2}{c}{\textbf{\underline{\qquad\quad 1-24h\quad\qquad}}} & \multicolumn{2}{c}{\textbf{\underline{\quad\qquad 25-48h\qquad\quad}}} & \multicolumn{2}{c}{\textbf{\underline{\quad\qquad 49-72h\qquad\quad}}} \\
&MAE&RMSE&MAE&RMSE&MAE&RMSE\\
\midrule
GCGRU\cite{li18} & 15.50 $\pm$ 0.11 & 18.40 $\pm$ 0.13 & 16.82 $\pm$ 0.10 & 19.80 $\pm$ 0.11 & 17.16 $\pm$ 0.11 & 20.14 $\pm$ 0.11\\
STGCN\cite{yu18} & 21.06 $\pm$ 0.20 & 24.65 $\pm$ 0.23 & 22.45 $\pm$ 0.18 & 26.16 $\pm$ 0.21 & 23.00 $\pm$ 0.13 & 26.72 $\pm$ 0.17\\
GWNET\cite{wu19} & 29.20 $\pm$ 0.21 & 33.97 $\pm$ 0.18 & 29.59 $\pm$ 0.17 & 34.45 $\pm$ 0.15 & 29.70 $\pm$ 0.12 & 34.54 $\pm$ 0.10\\
GCLSTM\cite{qi19} & 16.51 $\pm$ 0.53 & 19.50 $\pm$ 0.54 & 17.76 $\pm$ 0.51 & 20.80 $\pm$ 0.50 & 18.10 $\pm$ 0.41 & 21.14 $\pm$ 0.44\\
MSTGCN\cite{guo19} & 20.80 $\pm$ 0.12 & 24.36 $\pm$ 0.12 & 22.77 $\pm$ 0.12 & 26.43 $\pm$ 0.12 & 23.46 $\pm$ 0.19 & 27.17 $\pm$ 0.19\\
ASTGCN\cite{guo19} & 19.42 $\pm$ 0.13 & 22.83 $\pm$ 0.14 & 22.26 $\pm$ 0.10 & 25.84 $\pm$ 0.11 & 23.06 $\pm$ 0.10 & 26.71 $\pm$ 0.11\\
$\rm{PM_{2.5}GNN}$\cite{wang20} & 15.11 $\pm$ 0.11 & 17.94 $\pm$ 0.14 & 16.32 $\pm$ 0.13 & 19.23 $\pm$ 0.15 & 16.59 $\pm$ 0.16 & 19.51 $\pm$ 0.18\\
GAGNN\cite{chen21} & 19.54 $\pm$ 0.27 & 22.64 $\pm$ 0.24 & 20.71 $\pm$ 0.26 & 24.38 $\pm$ 0.26 & 21.24 $\pm$ 0.32 & 27.30 $\pm$ 0.29\\
HighAir\cite{xu21} & 15.93 $\pm$ 0.40 & 18.85 $\pm$ 0.42 & 17.07 $\pm$ 0.34 & 20.05 $\pm$ 0.35 & 17.36 $\pm$ 0.43 & 20.33 $\pm$ 0.45\\
\midrule
\textbf{M2G2} & \textbf{13.96 $\pm$ 0.09} & \textbf{16.78 $\pm$ 0.10} & \textbf{15.15 $\pm$ 0.08} & \textbf{17.91 $\pm$ 0.09} & \textbf{14.87 $\pm$ 0.09} & \textbf{17.60 $\pm$ 0.10}\\

\bottomrule
\label{baselineo3}
\end{tabular}}
\vspace{-1.em}
\end{table}

\section{Conclusion}

In this study, we introduce M2G2, a spatial-temporal dual multi-scale model that effectively captures complex relationships in spatiotemporal data at different scales and performs cross-scale fusion. Our proposed model leverages a bidirectional learnable fusion channel based on GCN to address the spatial dimension, allowing for effective utilization of multi-scale information. Additionally, we enhance the adaptive multi-scale updating mechanism based on GRU to handle the temporal dimension, dynamically adjusting the importance of different temporal-scale features in varying circumstances.

To evaluate the performance of M2G2, we collect a high-quality dataset encompassing a wide range of air pollutants and comprehensive meteorological indicators. On this real-world dataset, our model achieves optimal performance in predicting four types of air pollutants: PM2.5, PM10, NO2, and O3. Notably, M2G2 outperforms the second-best method in terms of MAE and RMSE metrics across three time periods: 1-24 hours, 25-48 hours, and 49-72 hours. The following outlines the improvements of M2G2 in comparison to the second-best method, based on the evaluation metrics of MAE and RMSE of the 24h/48h/72h: PM2.5: (6.22\%, 6.63\%, 9.71\%) and (7.72\%, 6.67\%, 10.45\%), ${\rm PM}_{10}$: (5.78\%, 5.52\%, 8.26\%) and (6.43\%, 5.68\%, 7.73\%), ${\rm NO}_2$: (5.40\%, 9.73\%, 19.45\%) and (5.07\%, 7.76\%, 16.60\%), ${\rm O}_3$: (7.61\%, 7.17\%, 10.37\%) and (6.46\%, 6.86\%, 9.79\%). These results effectively demonstrate the efficacy of M2G2 in capturing spatiotemporal multi-scale features of various air pollutants in real-world scenarios.

Furthermore, we observe that the improvements provided by M2G2 become more pronounced as the prediction time increases. This highlights the robustness of our approach in long-term prediction, as it exhibits less accuracy decay compared to short-term predictions. While the scales in this study are predetermined, future research can explore the design of dynamic spatiotemporal prediction models to further investigate the internal correlations within the data.

\section{Acknowledgement}

This work was supported by China Meteorological Administration Climate Change Special Program (CMA-CCSP) under Grant QBZ202316, the National Natural Science Foundation of China (Grant No. 62106116), the foundation of International Research Centre of Urban Energy Nexus, Hong Kong Polytechnic University (No. P0047700), Flexibility of Urban Energy Systems (FUES, No. P0043885) and Natural Science Foundation of Ningbo of China (No. 2023J027).

\end{spacing}

\bibliographystyle{unsrt}
\bibliography{References}
\end{document}